\documentclass[manuscript,screen]{acmart}

\usepackage{multirow}
\usepackage{tikz}
\usepackage{makecell}
\usepackage{threeparttable}
\usepackage{verbatim}
\usepackage{algorithm}
\usepackage{svg}
\usepackage{romannum}

\usepackage{algorithmic}
\usepackage{graphicx}
\usepackage{textcomp}
\usepackage{xcolor}

\newcommand{\IN}{\mathcal{X}}

\newcommand*{\circled}[1]{\lower.7ex\hbox{\tikz\draw (0pt, 0pt)%
             circle (.5em) node {\makebox[0em][c]{\small #1}};}}
\newcommand{\tabincell}[2]{\begin{tabular}{@{}#1@{}}#2\end{tabular}}
\newcommand{\commentout}[1]{}

\AtBeginDocument{%
  \providecommand\BibTeX{{%
    \normalfont B\kern-0.5em{\scshape i\kern-0.25em b}\kern-0.8em\TeX}}}

\setcopyright{acmcopyright}
\copyrightyear{2024}
\acmYear{2024}
\acmDOI{XXXXXXX.XXXXXXX}

\acmJournal{TOSEM}
\acmVolume{XX}
\acmNumber{Y}
\acmArticle{Z}
\acmMonth{8}

\acmPrice{15.00}
\acmISBN{978-1-4503-XXXX-X/18/06}

\begin{document}


\title{Anomaly Detection Based on Critical Paths for Deep Neural Networks}
\author{Fangzhen Zhao}
\affiliation{%
  \institution{School of Electronic \& Information Engineering (School of Big Data Science), Taizhou University}
  \city{Taizhou}
  \country{China}
}

\author{Chenyi Zhang}
\affiliation{%
  \institution{Computer Science and Software Engineering, The University of Canterbury}
  \country{New Zealand}
}

\author{Naipeng Dong}
\affiliation{%
  \institution{School of EECS, The University of Queensland}
  \country{Australia}}

\author{Ming Li}
\affiliation{%
  \institution{College of Information Science and Engineering, Jinan University}
  \city{Guangzhou}
  \country{China}
}

\author{Jinxiao Shan}
\affiliation{%
  \institution{Digital Center, China Merchants Group Limited}
  \city{Shenzhen}
  \country{China}
}

\begin{abstract}
Deep neural networks (DNNs) are notoriously hard to understand and difficult to defend. Extracting representative paths (including the neuron activation values and the connections between neurons) from DNNs using software engineering approaches has recently shown to be a promising approach in interpreting the decision making process of blackbox DNNs, as the extracted paths are often effective in capturing essential features. With this in mind, this work investigates a novel approach that extracts critical paths from DNNs and subsequently applies the extracted paths for the anomaly detection task, based on the observation that outliers and adversarial inputs do not usually induce the same activation pattern on those paths as normal (in-distribution) inputs.
 
In our approach, we first identify critical detection paths via genetic evolution and mutation. Since different paths in a DNN often capture different features for the same target class, we ensemble detection results from multiple paths by integrating random subspace sampling and a voting mechanism. Compared with state-of-the-art methods, our experimental results suggest that our method not only outperforms them, but it is also suitable for the detection of a broad range of anomaly types with high accuracy.
\end{abstract}



\maketitle

\section{Introduction} \label{sec:introduction}
Deep neural networks (DNNs) have gained a great advantage over traditional machine learning methods for image classification tasks in recent years~\cite{he2016deep,krizhevsky2012imagenet}.
However, it is well-known that DNN classifiers can be highly inaccurate (sometimes with high confidence) on test inputs from outside of the training distribution~\cite{nguyen2015deep,baseline,Mahalanobis}. Such inputs, named \emph{anomaly} samples, may arise in real-world settings either unintentionally due to external factors, or due to malicious adversaries that 
 induce prediction errors in the DNN and disrupt the system. This phenomenon raises doubts about the application of the DNN model to safety-critical systems.

To address this problem, many efforts have been made to detect anomaly samples in DNN models~\cite{meng2017magnet,InfluenceFunction,gong2019memorizing,kd+bu,LID,baseline,Mahalanobis,ma2019nic,hendrycks2018deep,sastry2020detecting,OODL}. Among them, methods based on the features 
extracted from one layer or several layers of a DNN model are widely used~\cite{kd+bu,LID,baseline,Mahalanobis,ma2019nic,hendrycks2018deep,sastry2020detecting,OODL} for their 
acceptable performance and easy-to-understand decision logic.
However, these methods do not usually generalize well across different types of anomaly samples because the activation values of a specific layer often depend on a specific anomaly type. Furthermore, an observation is that the features used in these methods are independent, meaning that the 
relationship of neurons between different layers in a DNN model 
is often ignored. While a DNN, which consists of neurons in multiple layers and connections among neurons in neighboring layers, can be considered as a weighted directed graph 
with nodes and weighted edges, and these connections form a huge set of paths from neurons of the input layer to neurons of the output layer. In the software engineering practice, it is well known that path-oriented testing is often beneficial for high-dependable testing performance of traditional software~\cite{wang2019deeppath}. Likewise, we believe that the connections between neurons of neighboring layers could provide meaningful characterization for a DNN model's functionality, and thus can be used as an indication for distinguishing normal 
and anomaly inputs.
 
 Researchers have already attempted to 
exploit the idea of \emph{critical path} borrowed
 from  the classical program analysis techniques~\cite{wang2018interpret,xie2022npc,li2021understanding,zhang2020dynamic,qiu2019adversarial}, and have used this concept both in understanding the decision logic of the DNN and for the detection of adversarial attacks. 
For the methods that use paths in a decision logic~\cite{wang2018interpret,xie2022npc,qiu2019adversarial,zhang2020dynamic} that is mostly 
 determined by the computational flows 
through neurons and connections, 
the generated \emph{critical decision paths} are usually sparse, and each path is defined for a unique decision class. 
 Similar techniques are also used for conveying both adversarial perturbations and important semantic information in a deep learning model~\cite{li2021understanding}. These paths, named as \emph{critical attacking paths}, can be 
useful in understanding model behaviors in adversarial attacks. 
 
Nevertheless, although the critical decision paths have been successfully used for detecting adversarial samples~\cite{wang2018interpret,zhang2020dynamic, zhang2020dynamic,qiu2019adversarial}, they have not been applied in detecting other types of anomaly samples such as out-of-distribution samples and random noises. When attempting to adopt existing critical paths extraction methods in detection of various anomaly types, we observe the following limitations.  
\begin{itemize}
\item 
Efficiency. The existing methods often require a significant amount of time for 
profiling a collection of paths
according to 
all inputs 
from a given training set, before applying an abstraction operation to merge these paths for each class and/or each adversarial attack type. 
\item Stablility. The existing methods often target on the decision making of a specific attack, which makes the detection performance unstable across other attack types and other anomaly~types. 
 \item Application Scope. The existing critical paths approaches have not been used to detect out-of-distribution samples and noise data. 
 \end{itemize} 

To address the limitation on efficiency, we propose a novel approach that 
extracts sequences of neurons, called \emph{critical detection paths}, from a
large volume of samples at once.
In our construction, a critical detection path starts from the input layer and ends at the output layer, 
which contains a single neuron for each layer. Instead of selecting one critical path for each sample, in our approach, we 
construct a mixed set 
from both normal and anomaly samples, and then use the mixed set to select one critical detection path for each class, inspired by the ways of genetic evolution~\cite{floreano2008bio} and mutation~\cite{ma2018deepmutation}. 
Different from previous approaches for critical decision paths and critical attacking paths~\cite{wang2018interpret,zhang2020dynamic, zhang2020dynamic,qiu2019adversarial}, our critical detection path has only one neuron for each layer, while the paths of other approaches may have several neurons for each layer. 
Moreover, our method generalizes well across various of anomaly samples, including different types of adversarial inputs, out-of-distribution inputs, and random noises, while the existing methods focus on detection and interpretation of adversarial inputs only. On the technical side, we extract several paths by using a random subspace algorithm~\cite{ho1998random}, and subsequently apply a voting mechanism that combines the results of several critical detection paths to achieve a better detection accuracy. 

In this paper, we propose 
a general purpose anomaly detection method 
called Anomaly Detection based on Critical Paths (ADCP). This approach is based on 
the observation that there is a significant difference between the distribution of normal and anomaly samples when using features on 
the extracted critical detection paths, as illustrated in Figure~\ref{fig:F-MNIST}. Therefore, we can predict whether an arbitrary 
input is anomaly or not by using the features represented as a vector of activation values 
spanning across the associated paths.

The contributions of this work can be summarised as follows:
\begin{itemize}
    \item[-] We propose a novel approach 
    that extracts critical detection paths from a DNN that are suitable for detecting adversarial (AD), out-of-distribution (OOD), and noise (NS) samples. Our critical detection paths contain a smaller number of neurons, as well as less connections between neurons.
    
    \item[-] Based on the detection results of several critical paths, we design  a novel detection method. This method improves detection accuracy and consistently 
   performs well for different types of anomaly samples 
    across various DNN  models and datasets.

    \item[-] We conduct 
    extensive experiments to show the effectiveness of the proposed approach, analyze the results, and demonstrate the importance of 
    critical detection paths in the detection process.        
\end{itemize}

\section{Problem Formulation} \label{sec:background}
A target deep neural network (DNN) model denoted as $\mathit{M}$, can be represented as a function $f:\IN\rightarrow C$ that maps each input $x \in \IN$ to a certain class $c \in C$. Assume $\mathit{M}$ contains $N$ 
layers (including the input layer, the hidden layers and the output layer), and the $i$-th 
layer is denoted as $L_{i}\ (0 \le i \le N-1)$. Assume there are $w_i$ number of neurons in the layer $L_{i}$, denoted as $n_{i}^{0}, n_{i}^{1}, \cdots, n_{i}^{w_i-1}$ respectively. Given a test input $x$, we denote the activation value of the neuron $n_{i}^{j}$ ($0 \le j \le w_i-1$) as $s_{i}^{j}(x)$: when layer $i$ is a fully connected layer, $s_{i}^{j}(x)$ is the activation value; when layer $i$ is a convolutional layer, then $s_{i}^{j}(x)$ is the average pooling value of channel (kernel) $j$.

\begin{definition}
A path of DNN model $M$ is defined as a sequence of neurons $\langle n_{0}^{v_0}, n_{1}^{v_1}, \ldots, n_{N-1}^{v_{N-1}} \rangle$ where 
\begin{itemize}
    \item 
    neuron $n_{i}^{v_i}$ is at the $i$-th layer $L_i$, with $0\leq v_i\leq w_i-1$,
    \item neighboring neurons belong to two connected layers, i.e., if $n_{i}^{v_i}$ is followed by $n_{j}^{v_j}$ in the path, $j=i+1$.
\end{itemize}
A path is instantiated with values when a test input $x$ is given, formally, the instance of a path $\langle n_{0}^{v_0}, n_{1}^{v_1}, \ldots, n_{N-1}^{v_{N-1}} \rangle$ with a given input $x$ is $\langle s_{0}^{v_0}(x), \ldots, s_{N-1}^{v_{N-1}}(x)\rangle$.
\end{definition}

Therefore, selecting one neuron from each layer following the order of layers will form a path. 
For an $N$-layer model, the length of the path is $N$, and in-total there are $w_0*w_1* \ldots * w_{N-1}$ number of paths,  which is often a large number. This work aims to select a small subset of paths that could be effectively used for the purpose of anomaly detection.

Essentially the function $f$ of a model $M$ is determined based on a set of training data---a set of input $x$ each of which has a label $c_{x}$ to denote its ground-truth class. The DNN model generally obtains a generalization area of input based on the training data. If an input $x$ is outside the generalization area, we usually take it as an anomaly
~\cite{lust2020survey}. Three types of anomaly are considered in the literature depending on how they are generated: Adversarial (AD) data is generated by adding perturbation to a clean input; Out-of-Distribution (OOD) data refers to an input that are sampled from a different distribution of the training data; Noise (NS) data are random noise (e.g. Gaussian noise) and fooling images. The fooling images are created by evolving meaningless images in order to mislead a DNN to output classes with high confidence~\cite{nguyen2015deep}. As the literature suggests that the characteristics of each class can be captured by different set of paths, we conjunct that anomaly data requires a specific set of paths to represent, and these set of paths can be used for anomaly detection.

\paragraph{Support Vector Domain Description.}
Our anomaly detection algorithm is based on one class classification, 
which has been well studied for anomaly detection tasks~\cite{ma2019nic,deep-svdd(ICML2018),ruff2020deep}. In our work, we use the one class classification known as Support Vector Domain Description (SVDD) by~\cite{Tax2004supportvector}. Another important approach for one class classification known as $\nu$-SVC is introduced by Sch\"{o}lkopf et al.~\cite{schoelkopf2001}, which can be shown as equivalent to SVDD when the Gaussian kernel is used~\cite{tax:thesis}. Similar to the famous Support Vector Machine~\cite{vapnik1995}, SVDD defines support vectors for a sphere shaped decision boundary enclosing the class of objects represented by the (unlabeled) training data with minimal space, as shown in the following formulation. 

\begin{equation}\label{eq:svdd}
     \min\limits_{R,a,\xi}\ R^2+ \frac{1}{n\nu}\sum_i\xi_i
\end{equation}
\[\mbox{s.t. }\forall i:\quad \|x_i-a\|^2\leq R^2+\xi_i,\quad \xi_i\geq 0\]

The solution of the above constraints provides a center vector $a$, radius $R$ and slack variables $\xi_i$ such that the target term in Eq.~(\ref{eq:svdd}) is minimized, provided that the square of distance from each training data $x_i$ to the center $a$  may exceed $R^2$ by at most $\xi_i$. Here $\nu$ is a constant in $(0,1]$ and $n$ is the size of the training set. Intuitively, a smaller $\nu$ gives more weight to the right hand side of target term in Eq.~(\ref{eq:svdd}), which imposes smaller values for $\xi_i$ and larger $R$. The solution of Eq.~(\ref{eq:svdd}) allows us to determine if a test input $z$ is from the normal data by checking the following condition.
\begin{equation}\label{eq:if-id}
  \|z-a\|^2=(z\cdot z)-2\sum_i\alpha_i(z \cdot x_i)+\sum_{i,j}\alpha_i\alpha_j(z \cdot x_i)\leq R^2
\end{equation}
Here $\alpha_i$ ($\alpha_j$) is the Lagrange multiplier associated with the constraint for the $i$-th ($j$-th) training input when solving Eq.~(\ref{eq:svdd}), which is non-zero only if the $i$-th ($j$-th) training input is used as a support vector. Given all inputs (including the test input) only appearing in the form of inner product, it is thus viable to replace the inner products by kernel functions, of which the Gaussian Radial Basis Function (RBF) provides the best performance in practice~\cite{tax:thesis}.
The RBF kernel is given in the following formulation, where the free parameter $s$ controls the spread, or how tight the density is, of the kernel.
\begin{equation}\label{eq:rbf}
  K(x_{i}, x_{j}) = \exp(-\|x_{i} - x_{j}\|^{2}/s^2)
\end{equation}

During production runs, we only use the training set to train SVDD and compute the probabilities for the extracted critical paths. In other words, we collect the probabilities with SVDD models across all critical paths. Given an input $x$, a standard SVDD classifier is used to calculate the probability for each critical path. We can conject whether $x$ is an anomaly sample based on these probabilities and report the results. Details are illustrated in Subsection~\ref{approach:detection}.

\section{Our Approach} \label{sec:our approach}
\begin{figure}[t] 
  \centering
  \includegraphics[scale=0.6]{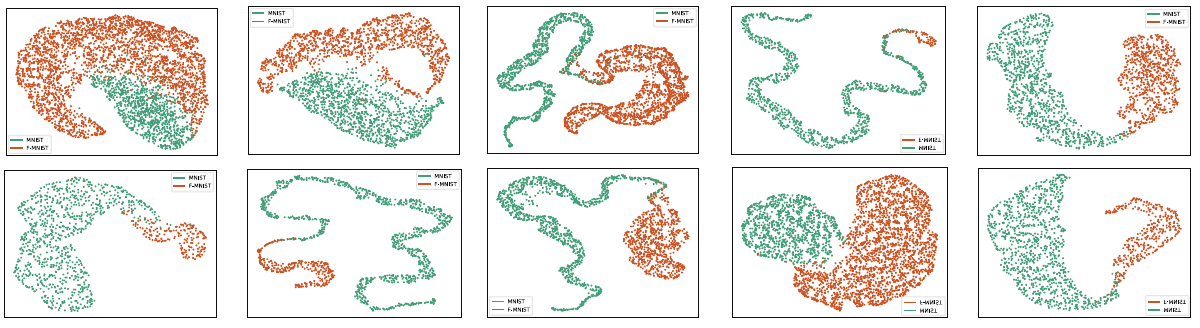}
  \caption{Two-dimensional representations of features extracted from the critical detection path of a LeNet model trained on MNIST. The feature clusters for the $10$ classes are shown, with green dots for MNIST (normal) data and brown dots for F-MNIST (anomaly) data.}\label{fig:F-MNIST}
\end{figure}

In this section, we present the methodology known as Anomaly Detection based on Critical Paths (ADCP).
We have conducted a preliminary study taking a LeNet model with the MNIST dataset as training (i.e., in-distribution) data, and we treat the Fashion-MNIST dataset as anomaly (out-of-distribution) data. The extracted paths features are used to separate the two datasets, with the results visualized in Figure~\ref{fig:F-MNIST}.
There are $10$ figures corresponding to the $10$ MNIST classes, due to that we perform the separation based on the output of the LeNet model. For example, the upper-left figure displays all data classified as class $1$ by the LeNet model, of which MNIST data are depicted in the green color and Fashion-MNIST data are all depicted in the brown color. These figures have indicated a 
reasonable separation of anomaly data from 
in-distribution samples for all of the $10$ MNIST classes. 

Following this observation, we propose 
our approach in the following two phases: 
path extraction and anomaly detection.

\begin{figure}[t] 
  \centering
  \includegraphics[scale=0.4]{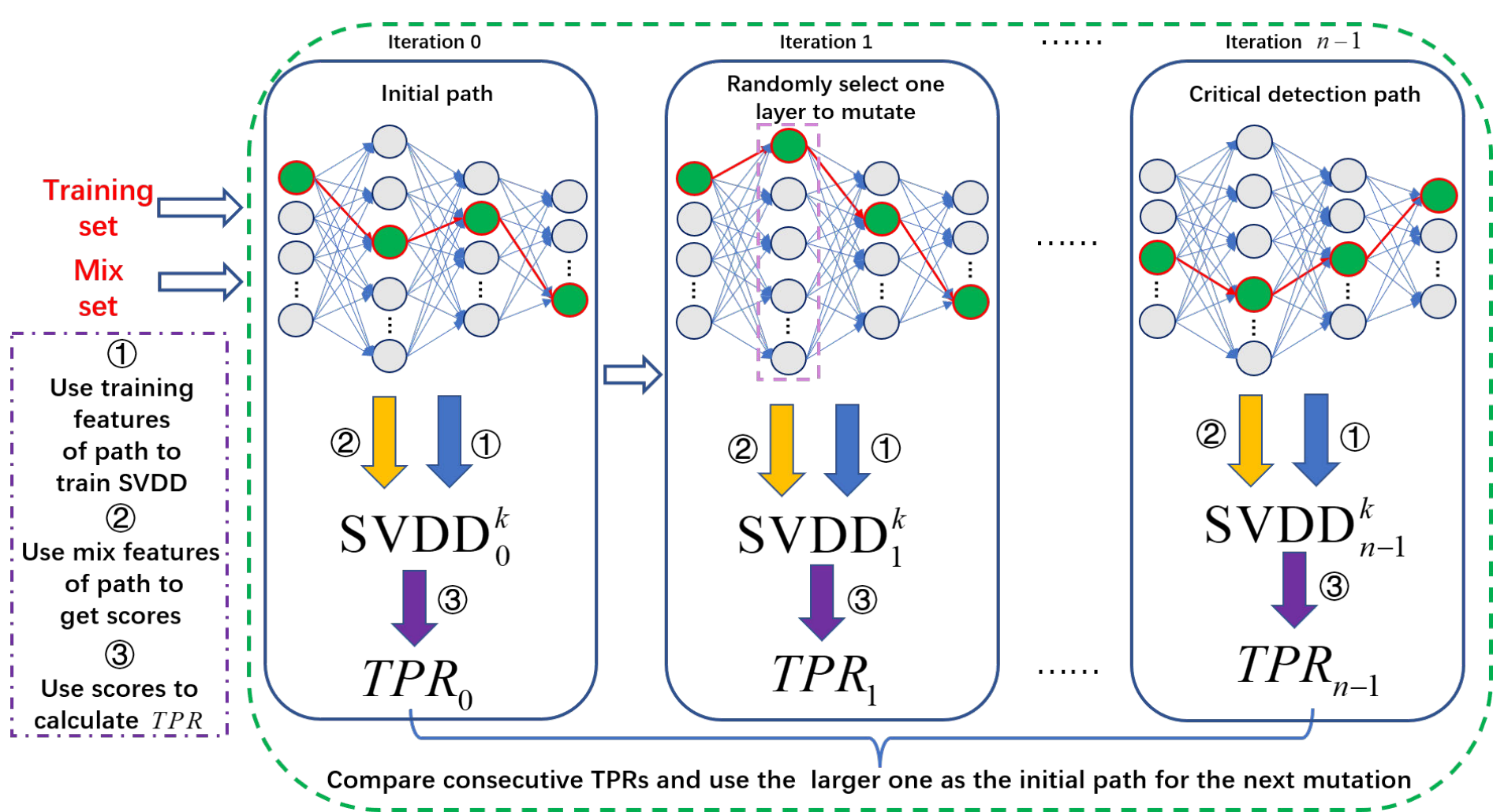}
  \caption{The overview of our approach for selecting critical detection path.}\label{fig:overview_paths}
\end{figure}

\subsection{Critical Detection Path Extraction} \label{approach:extraction}
We propose to use genetic evolution~\cite{floreano2008bio} and mutation~\cite{ma2018deepmutation} to extract a set of paths that are used in the subsequent analysis. The entire process is shown in Figure~\ref{fig:overview_paths}, and we define a selection metric to evaluate performance of the paths, which is given in the next paragraph. At the beginning, an initial path is randomly selected. Then in the next iteration, we randomly select one layer and mutate the corresponding neuron (by randomly selecting a different neuron of the same layer). The scores before and after the mutation are compared, and if the score of the mutated path is higher, then the old path is (greedily) replaced by the new path; otherwise the old path is retained. In this way, we evolve the critical path by mutating one neuron at a time until a given number of iterations are completed.

To calculate the selection metric, each iteration consists of three procedures (see Figure~\ref{fig:overview_paths}): $\circled{1}$ We use the features (neuron activation values as detailed later) of the training set on a given path of neurons
to train a Support Vector Domain Description (SVDD) model. $\circled{2}$ We feed the features of a mixed set of normal and anomaly samples on the same path to the pretrained SVDD model to obtain a score for each sample. $\circled{3}$ We calculate the selection score for this path by integrating these individual scores. This is implemented by calculating the True Positive Positive Rate (TPR) which is often used as a metric to evaluate the performance. 

\begin{equation}\label{eq:TPR}
    TPR = \frac{TP }{TP + FN} 
\end{equation}

where true positive (TP)  is the number of cases when an anomaly sample is correctly reported as anomaly, and false negative (FN) is the number of cases when an anomaly is incorrectly reported as normal. The TP and FN are obtained from the individual scores by comparing with a given threshold $\tau$: for an anomaly data $x$, if its score is below the corresponding threshold $\tau$, it is reported as TP, otherwise, it is reported as FN. The threshold is different for each class, and is calculated in a way to ensure that $95\%$ of the normal samples from this class have scores above $\tau$. 

As discussed in the literature, the characteristics for each decision class are different, and thus the critical detection path is different for each class. Since there is randomness in the path extraction process i.e., we randomly select the initial path, randomly select a layer to mutate, and randomly select a neuron in the layer, selecting only one path as the result may miss out other important paths. Therefore, we select multiple paths as the critical detection paths for each class (see the evaluation section for a discussion on the optimal number of paths). 
Since we use the same procedure to extract a critical detection path for each class, at this point, we focus on the path generation process assuming a particular class $k$ in the following paragraphs.

We first randomly select a neuron from each hidden layer to form an initial path. Then we feed all samples belonging to class $k$ from the training set into the model $\mathit{M}$ and obtain the neuron activation values for the selected neurons from these input samples. If the hidden layer is a convolutional layer, the neuron activation value is the average pooling value corresponding to the selected channel.

Next, we construct a mixed set consisting of anomaly and normal samples belonging to class $k$. The normal samples are from a test set. Note that we use different anomaly samples to form the mixed set for different types of anomaly detection tasks. In particular, for AD detection, we consider two scenarios. 
In one scenario, anomaly samples are a collection of adversarial samples of different attack methods. Taking into account the number of normal samples on the test set, for MNIST and CIFAR-10, we randomly select $200$ adversarial samples for each class from each attack, and for SVHN, we select 400 adversarial samples for each class from each attack. Therefore, the number of normal samples and anomaly samples is relatively close among the mixed set.
In the other scenario, anomaly samples only consist of all FGSM inputs. For OOD/NS detection, we pick one OOD set as anomaly sample. For MNIST, we pick F-MNIST as anomaly set. For CIFAR-10 and SVHN, we pick TinyImageNet as anomaly set.  We feed all mixed samples into $\mathit{M}$ and obtain the neuron activation values for the selected neurons from these mixed samples. Suppose the mixed set consists of $2000$ samples, the features of an initial path is a matrix of size $2000*5$ (suppose the target DNN has $5$ layers), denoted as $V_{0}^{*}$. We feed $V_{0}^{*}$ into SVDD$_{0}^{k}$, and SVDD$_{0}^{k}$ returns a score for each sample (corresponding to \circled{2} in Figure~\ref{fig:overview_paths}). Then we use these scores to calculate the True Positive Rate (TPR), donated as TPR$_{0}$ (corresponding to \circled{3} in Figure~\ref{fig:overview_paths}). 

In the second iteration, we randomly select one layer to mutate, and the unselected layers remain unchanged.  
Taking the LeNet (for MNIST) as an example, assume the initial path is $\langle n_{0, 1}, n_{1, 3}, n_{2, \textbf{5}}, n_{3, 7}, n_{4, 9}\rangle$. We select the 2-nd layer to mutate, the path then becomes $\langle n_{0, 1}, n_{1, 3}, n_{2, \textbf{8}}, n_{3, 7}, n_{4, 9}\rangle$. By repeating the steps $\circled{1}$ and $\circled{2}$ for the mutated path in the second iteration, we can get a vector $V_{1}$ and $V_{1}^{*}$ from the training set and mixed set in separate. We use $V_{1}$ to train another SVDD model SVDD$_{1}^{k}$ and feed $V_{1}^{*}$ into SVDD$_{1}^{k}$, which returns a score for each sample. We use these scores to calculate another TPR, denoted as TPR$_{1}$. Then we compare TPR$_{0}$ and TPR$_{1}$, and take path with larger score as seed for the next iteration. Note that if the selected layer has just been mutated previously, we skip this iteration. Finally, a path $p_{i}$ with the highest TPR is returned as a critical detection path.

The entire approach is formalised in Algorithm~\ref{alg:1}, where the operator $Mutate(p_i,l)$ returns a path information with higher TPR for the selected path $p_i$ and the selected mutated layer $l$.

\begin{algorithm}[h]
\small
	\caption{Get critical detection paths}
	\label{alg:1}
	\begin{algorithmic}[1]
	\REQUIRE~~   \\ 
	$\mathit{M}$: DNN model; \\
	X$_{train}$: all samples belonging to class $k$ from the training set; \\
	X$_{mix}$: all samples belonging to class $k$ from the mixed set; \\
	$m$: the number of critical detection paths; \\
	$n$: the number of mutations; \\
    \ENSURE~~   \\ 
     P: $m$ critical detection paths  \\
     \STATE P = \{\}
	\FOR{$j = 0, \cdots, m-1$}
	  \STATE  Randomly selected one initial path $p_0$ 
        \STATE  Calculate $TPR_0$ 
	  \STATE LastVisit := -1
	  \FOR{$i = 0,\cdots,n-1$}
	  \STATE  Randomly selected one layer $l$ to mutate
	  \IF{LastVisit $\ne l$}
	  \STATE  $p_{i} \gets  Mutate(p_0,l)$ 
        \STATE  Calculate $TPR_i$ 
	  \STATE LastVisit $ = l$
	  \ENDIF
        \IF{$TPR_i > TPR_0$}
	    \STATE $p_{0} = p_{i}$
            \STATE $TPR_{0} = TPR_{i}$
        \ENDIF
	  \ENDFOR
	  \STATE P $\gets $ [P; $p_{0}$]
	  \ENDFOR
	\RETURN P
	\end{algorithmic}
\end{algorithm}

\subsection{Anomaly Detection} \label{approach:detection}
Based on the selected critical detection paths, we design a novel approach to detect anomaly samples 
by using the random subspace algorithm~\cite{ho1998random} and a voting mechanism. 

First, we sort the critical detection paths 
represented as a sequence of $TPR$s in descending order. Then we feed all the samples of each class in the training set into $\mathit{M}$ in separate, and extract the features of the corresponding critical detection paths. Taking class $k$ as an example, if there are $m$ critical detection paths, we can get $m$ vectors, denoted as $V_{1}, V_{2}, \cdots, V_{m}$. Then we use each $V_{i} 
 \ (1\le i\le m) $ to train an SVDD model and we finally obtain $m$ SVDD models for class $k$, denoted as SVDD$_{1}$, SVDD$_{2}, \cdots, $ SVDD$_{m}$. Taking the LeNet (for MNIST) as an example, in-total, we get $m \times 10$ SVDD models ($10$ classes), denoted as SVDD$_{i}^{j}$, where $ 
 1\le i\le m$ and $0\le j \le 9$. These SVDD models are used in the following anomaly detection procedure.

Given a test data, first, we run $\mathit{M}$ to obtain a predicted class $k$. Then, we extract the features of the corresponding critical detection paths as $V_{i}^{'}\ (1\le i\le m)$. $V_{i}^{'}$ is forwarded to the pretrained model SVDD$_{i}^{k}$ 
to get a score. For $m$ critical detection paths, we obtain $m$ scores. As different critical paths tend to generate scores at different scales, we apply the min-max normalization procedure. 
\begin{equation}\label{Eq:nomalization}
  score_{i}^{*} = \frac{score_{i} - score_{min}}{score_{max} - score_{min}}
\end{equation}
where the $score_{min}$ and $score_{max}$ are the minimum and maximum of the score vector, respectively.

For each critical detection path, we set a threshold $\tau$ for each class in advance. The threshold is the same as described in subsection~\ref{approach:extraction}. We initialize two sets $A$ and $B$ to save scores in separate. If the score is above the threshold, the score is saved to $A$, and if it is below the threshold, the score is saved to $B$. The final score is calculated using a voting mechanism based on $A$ and $B$ as follows:
\begin{itemize}
    \item If there are $m$ values in $A$, it means that the given test data is decided as a normal sample for all critical detection paths. The final score is the maximum score among all the scores in $A$.
    \item If there are $m$ values in $B$, it means that the test data is decided as an anomaly sample for all critical detection paths. The final score is the minimum value in $B$.
    \item If there are $m^{'}$ 
    values in $A$ and $m-m^{'}$ values in $B$, we use a voting mechanism to decide the score. If $m^{'} > m-m^{'}$, we take the median value in $A$ as the final score, otherwise, we take the median value in $B$. 
\end{itemize}

\paragraph{Threshold.} Once a final score for the test data is obtained, we use it to decide whether the input is normal or anomaly. Similar to  existing methods~\cite{baseline,ODIN,GeneralizedODIN}, we achieve this by setting thresholds. Different from most other works, we define multiple thresholds based on classes of training samples. In the case of MNIST, there are $10$ classes of training data, so we define a threshold for each class. When a sample $x$ is given to the target DNN model which generates output class $i$, our approach collects features from each critical path for the corresponding SVDD to generate score. Then we utilize the scores of $21$ critical paths to get the final $score$ to compare with $\tau_i$. The threshold $\tau_{i}$ is computed in the same way while it need ensure that $95\%$ of the normal samples from class $i$ of test sets have scores above $\tau_{i}$. The threshold-based discriminator can be formally described as follows.

\begin{equation}\label{threshold}
    isAnomaly(x) = \left\{
    \begin{aligned}
     & True, \;\;\; if \;\; score <  \tau_{i} \\
     & Fasle, \;\;\; if \;\; score \geq \tau_{i}
    \end{aligned}
    \right.
\end{equation}

\section{Evaluations} \label{sec:evaluations}
We aim to investigate and answer the following research questions through our experiments: 

\begin{itemize}
  \item \textbf{RQ1:} Does Algorithm~\ref{alg:1} generate critical paths that have significantly better performance than initial paths?  
    \item \textbf{RQ2:} How effective is ADCP compared with state-of-the-art 
    anomaly detection methods?   
    \item \textbf{RQ3:} Does ADCP have better performance in anomaly detection than the existing path-based approaches?
    \item \textbf{RQ4:} How does ADCP perform in terms of different hyper-parameters?   
\end{itemize}
\subsection{Experiment Setup} 
\emph{Hardware and software.} We conducted our experiments on two different platforms. One is Windows $10$ desktop equipped with Intel I7-9700 $3.0$GHz processor, $16$G RAM and Nvidia GetForce GTX1660Ti. The other one is Linux equipped with Intel(R) Xeon(R) Gold 6230 $2.10$GHz processor, $64$G RAM and Nvidia Titan RTX 24G. We have implemented the experiments based on the Keras framework. We conduct experiment on three types of target DNN models, with three data sets chosen as normal samples, against various types of AD, OOD and NS data sets. Our testing code is publicly available at \url{https://github.com/fangzhenzhao/ADCP}.

\emph{Datasets and models.}
We evaluated our approach (ADCP) on the following well-known image classification datasets and models (see Table~\ref{table:models}). 
\begin{table}[h]
\caption{The datasets and models of our work.}\label{table:models}
\centering
\begin{tabular}{lllll}
\cline{1-3}
\multicolumn{1}{c}{Dataset}                   & \multicolumn{1}{c}{Model}    & \multicolumn{1}{c}{Accuracy  on Test Dataset} &  &  \\ \cline{1-3}
\multicolumn{1}{c}{MNIST}                     & \multicolumn{1}{c}{LeNet}    & \multicolumn{1}{c}{99.20\%}                    &  &  \\ \cline{1-3}
\multicolumn{1}{c}{\multirow{2}{*}{CIFAR-10}} & \multicolumn{1}{c}{VGG16}    & \multicolumn{1}{c}{93.47\%}                    &  &  \\ \cline{2-3}
\multicolumn{1}{c}{}                          & \multicolumn{1}{c}{ResNet44} & \multicolumn{1}{c}{91.65\%}                    &  &  \\ \cline{1-3}
\multicolumn{1}{c}{\multirow{2}{*}{SVHN}}     & \multicolumn{1}{c}{VGG16}    & \multicolumn{1}{c}{95.56\%}                    &  &  \\ \cline{2-3}
\multicolumn{1}{c}{}                          & \multicolumn{1}{c}{ResNet20} & \multicolumn{1}{c}{96.12\%}                    &  &  \\ \cline{1-3}
\end{tabular}
\end{table}

\emph{Anomaly samples.} We generate three types of anomaly: 
\begin{itemize}
\item 
AD: We generate the adversarial attacks by using the open-source CleverHans library \cite{papernot2018cleverhans} on TensorFlow. The adversarial attacks used in our experiments include FGSM~\cite{FGSM},  PGD~\cite{meng2017magnet}, JSMA~\cite{PapernotMJFCS15}, DeepFool~\cite{moosavi2016deepfool} and CW~\cite{cw}.
\item
OOD: For MNIST, the OOD sets are \textbf{Fashion-MNIST (F-MNIST)}~\cite{Fashion} and \textbf{Omniglot}~\cite{Omniglot}. For CIFAR-10, the OOD sets are \textbf{TinyImageNet}~\cite{Imagenet}, \textbf{LSUN}~\cite{LSUN}, \textbf{iSUN}~\cite{iSUN} and \textbf{SVHN}~\cite{svhn}. For SVHN, the OOD sets are \textbf{TinyImageNet}, \textbf{LSUN}, \textbf{iSUN} and \textbf{CIFAR-10}. 
\item
NS: We prepare three types of noise images, namely Gaussian Noise, Uniform Noise and Fooling Image~\cite{nguyen2015deep}.
\end{itemize}

\emph{Metrics.} 
We adopt two commonly used metrics, True Negative Rate at $95\%$ of True Positive Rate (TPR at $95\%$ TNR) and Area Under the Receiver Operating Characteristic curve (AUROC), to evaluate the effectiveness of our method.

\subsection{RQ1: Comparison with Initial Paths}
\begin{figure}[h]
  \centering
  \includegraphics[scale=0.62]{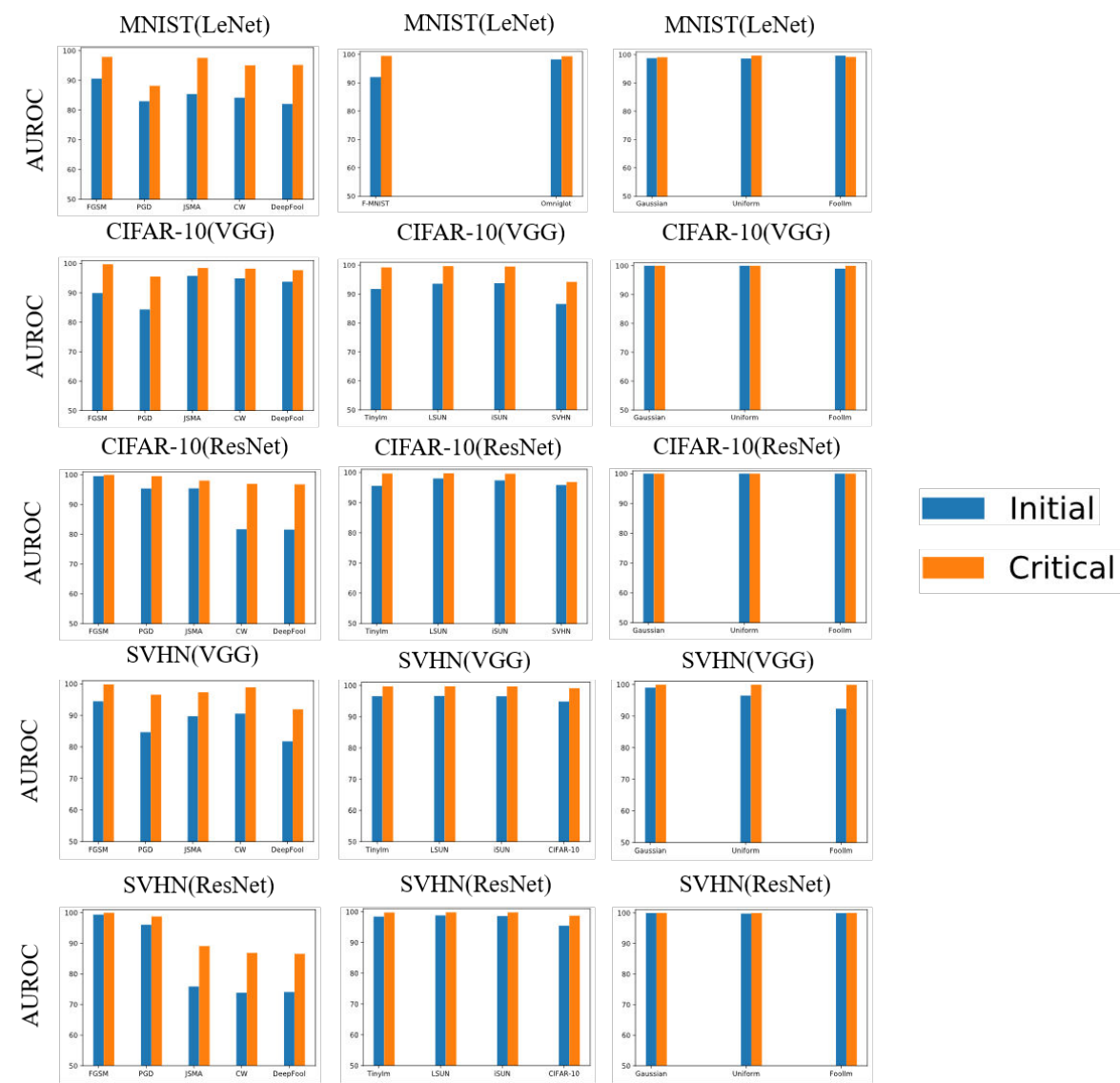}
  \caption{The AUROC corresponding to different paths on five models for different anomaly samples.}\label{fig:Different_paths}
\end{figure}

We answer \textbf{RQ1} by comparing the AUROC of different paths. A highest TPR among selected $m$ initial path (Initial in Figure~\ref{fig:Different_paths}), 
and a path with the highest TPR among the selected $m$ critical detection paths---our approach (Critical in Figure~\ref{fig:Different_paths}) are applied to different types of anomaly samples. The experimental results are shown in Figure~\ref{fig:Different_paths}.

It can be seen from Figure~\ref{fig:Different_paths} that the AUROC of the Critical path are significantly better than the Initial path  for AD and OOD detection tasks. For the NS detection task, the advantage of the critical paths is not obvious in some cases, e.g. the difference is insignificant for Gaussian noises and Uniform noises 
in all models with CIFAR-10 and SVHN as in-distribution data. 

\begin{center}
\fcolorbox{black}{gray!10}{\parbox{.95\linewidth}{\textbf{Answer to RQ1:} Our critical detection path is more effective than the initial path (highest TPR among the initial paths $m$)  for detecting various types of anomaly samples.}}
\end{center}

\subsection{RQ2: Comparison with Other Anomaly Detection Methods}
We answer \textbf{RQ2} by comparing the performance of our approach with the traditional adversarial detection approaches.  

\paragraph{Baseline Methods} Existing amonaly detection works mostly focus on detecting either OOD or AD only. Therefore, we compare our results on OOD and NS data with models designed for OOD detection, and compare our results on AD data with models designed for adversarial attack detection, in separate. For AD detection, the baseline methods are: Local Intrinsic Dimensionality (LID)~\cite{LID},  Mahalanobis Distance (MD)~\cite{Mahalanobis},  MagNet~\cite{meng2017magnet}, Feature Squeezing (FS)~\cite{FeatureSqueezing}, Neural-network Invariant Checking (NIC)~\cite{ma2019nic}, Anomaly Detection based on Critical Layers (ADCL)~\cite{zhao2022uniform}. For OOD/NS detection, the baselines are: Memory AutoEncoder (MemAE)~\cite{gong2019memorizing},  Max-Softmax (MS)~\cite{baseline}, Out-of-DIstribution detector for Neural networks (ODIN)~\cite{ODIN}, Mahalanobis Distance (MD)~\cite{Mahalanobis}, Early Layer Output (ELO)~\cite{OODL}, and Anomaly Detection based on Critical Layers (ADCL)~\cite{zhao2022uniform}. These methods are detailed in Section~\ref{sec:related works}.

\paragraph{AD Detection} Among existing detecting AD data approaches, LID and MD need adversarial samples to train the detectors. However, in practice, we do not usually know whether the test data is clean or adversarial, let alone what kind of adversarial sample it is. Therefore, for a valid comparison, we make the following settings for LID and MD.  For MD, we do not test MD with feature ensemble which uses the outputs from all layers and involves tuning with particular anomaly samples. We only apply the version of MD which uses the logit layer of $\mathit{M}$ instead, which have been used in the literature \cite{ren2019likelihood,OODL}. For LID, we use two different adversarial samples for training. The first is by only using FGSM sample as adversarial inputs, the second is by a sample mixed with all different types of adversarial inputs used in this paper, with resulting detectors called LID$^{*}$ and (non-annotated) LID, respectively. For fair comparison between LID and ADCP, we also use two types of adversarial samples to train ADCP$^{*}$and ADCP, so that their results are compared with their corresponding  LID$^{*}$ and LID detectors, with results shown in Table~\ref{table:attack results AD}.

\begin{table}[H] 
    \caption{Comparison of our approach with LID, MD, MagNet, FS and  NIC for different attack on various models and datasets. Top 2 are highlighted with \textbf{bold} and \underline{underline}.}\label{table:attack results AD}
    \small
    \centering
    \begin{tabular}{@{\hskip -0.01cm}c@{\hskip -0.01cm}ccc}
        \toprule
         \multirow{2}{*}{Model} & \multirow{2}{*}{Attack}    & TPR at 95\% TNR $\uparrow$    & AUROC $\uparrow$ \\ \cline{3-4}
          &  & \multicolumn{2}{c}{LID$^{*}$ / LID / MD / MagNet /  FS / NIC / ADCP$^{*}$ / ADCP }  \\
    \midrule
        \multirow{3}{*}{{\tabincell{c}{LeNet\\ (MNIST)} } }    
         & FGSM    &  64.98/59.51/83.85/30.23/\underline{98.67}/49.68/\textbf{99.96}/93.99    & 94.12/93.29/96.82/88.56/\underline{99.55}/88.03/\textbf{99.93}/98.99   \\ [-3pt]
        & PGD   &  30.38/38.43/20.82/\;8.46/\textbf{99.67}/15.81/65.09/\underline{76.19}     & 83.87/85.82/77.71/66.38/\textbf{99.92}/71.06/91.98/\underline{97.76}   \\ [-3pt]
        &JSMA    & 80.91/86.35/89.95/24.09/\textbf{99.79}/70.18/89.85/\underline{94.12}  &
        96.54/97.47/97.82/82.78/\textbf{99.33}/93.87/98.08/\underline{98.81} \\ [-3pt]
        & CW  & 67.64/72.89/66.33/\;5.54/\textbf{100.0}/30.42/57.47/\underline{82.82}    & 94.20/95.75/95.12/60.66/\textbf{99.36}/84.20/93.80/\underline{97.24}   \\  [-3pt]
        & DeepFool    & 62.57/70.55/66.58/\;5.69/\textbf{100.0}/33.05/60.63/\underline{85.43}  & 93.75/95.50/95.29/62.40/\textbf{99.47}/86.95/93.96/\underline{97.44}   \\ 
         
    \midrule
     
    \multirow{3}{*}{{\tabincell{c}{VGG\\ (CIFAR-10)} } }  
        & FGSM    &  44.04/\;8.95/53.08/\;5.03/\;1.49/53.53/\textbf{100.0}/\underline{99.65}   & 88.30/70.20/\underline{90.07}/52.18/64.18/87.26/\textbf{99.96}/\underline{99.56}   \\ [-3pt]
         & PGD   & 11.56/15.72/\;5.83/\;4.61/14.60/20.15/\underline{72.11}/\textbf{82.34}     & 66.97/67.71/80.40/49.78/63.11/73.24/\underline{94.54}/\textbf{96.57} \\ [-3pt]    
    &JSMA    & 17.17/60.73/\underline{97.13}/\;5.24/\;0.17/71.32/92.44/\textbf{97.46}  & 69.04/93.81/\underline{98.78}/51.21/63.96/93.56/98.52/\textbf{99.15}\\  [-3pt]
    & CW  & \;0.0\; /45.66/\underline{86.22}/\;6.27/\;0.0\; /82.60/83.35/\textbf{94.92}     & 29.41/88.60/\underline{96.96}/54.17/51.59/96.30/96.68/\textbf{98.97} \\  [-3pt]
    & DeepFool    &  \;0.0\; /46.87/81.07/\;6.43/\;0.0\; /\underline{82.04}/79.23/\textbf{92.42}   & 30.01/87.77/\underline{96.18}/54.30/51.43/\underline{96.18}/95.86/\textbf{98.54} \\   

     \midrule
     
     \multirow{3}{*}{{\tabincell{c}{ResNet\\ (CIFAR-10)} } }  
         & FGSM    &  \underline{99.94}/70.33/37.69/\;4.40/\;4.53/48.68/\textbf{100.0}/\textbf{100.0}   & \textbf{99.97}/93.33/87.33/50.31/70.61/86.47/\underline{99.95}/99.93  \\  [-3pt]
 & PGD   & \;0.0\; /70.01/85.07/\;4.42/14.18/24.07/\underline{93.38}/\textbf{99.53}     & 47.81/92.41/95.87/49.43/59.83/71.55/\underline{98.40}/\textbf{99.76}   \\ [-3pt]
     &JSMA    & 14.86/70.01/35.25/\;9.78/\;1.16/69.94/\underline{88.80}/\textbf{94.15}   & 62.23/92.57/92.18/62.47/71.87/93.04/\underline{96.79}/\textbf{98.42}  \\  [-3pt]
     & CW  & \;0.0\; /\underline{57.36}/\;9.74/\;6.69/\;0.0\; /32.86/30.86/\textbf{85.04}     & 38.17/\underline{92.85}/82.99/54.67/52.88/78.42/84.91/\textbf{97.63}  \\  [-3pt]
    & DeepFool    & \;0.0\; /\underline{61.70}/\;8.74/\;6.55/\;0.0\; /32.70/30.62/\textbf{84.66}   & 38.13/\underline{93.04}/82.46/54.63/53.89/77.89/84.43/\textbf{97.53}  \\  
    
     \midrule
     
     \multirow{3}{*}{{\tabincell{c}{VGG\\ (SVHN)} } }  
         & FGSM    &  \underline{97.57}/77.28/81.58/76.52/75.67/80.60/\textbf{100.0}/\textbf{100.0}   & 99.41/96.11/96.49/97.36/95.82/95.99/\textbf{99.98}/\underline{99.93}   \\  [-3pt]
      & PGD    &  24.99/34.17/\;2.96/58.38/\;2.79/30.21/\underline{85.78}/\textbf{89.75}    & 81.52/82.32/80.03/92.17/37.52/78.16/\underline{97.16}/\textbf{97.88}  \\ [-3pt]
     &JSMA    & 36.51/72.29/93.81/70.79/\textbf{96.46}/49.08/81.98/\underline{95.34}   & 85.88/95.33/98.52/94.53/\textbf{99.07}/85.40/96.36/\underline{98.67} \\  [-3pt]
     & CW  & 16.16/73.34/\textbf{100.0}/19.91/98.68/24.05/88.02/\underline{99.70}     & 77.99/95.40/98.48/81.19/\underline{99.04}/72.97/97.88/\textbf{99.55}  \\  [-3pt]
     & DeepFool    &  27.60/55.67/67.90/42.97/\underline{75.07}/30.67/61.79/\textbf{81.50}    & 79.64/90.65/92.59/87.96/\textbf{95.05}/78.23/89.85/\underline{94.48}   \\  

     \midrule
     
    \multirow{3}{*}{{\tabincell{c}{ResNet\\ (SVHN)} } }  
        & FGSM    &  \underline{98.74}/75.89/68.41/77.18/83.91/54.83/\textbf{100.0}/\textbf{100.0}  & 99.40/95.67/94.71/98.55/97.14/89.84/\textbf{99.98}/\underline{99.90}   \\  [-3pt]
     & PGD    & 61.97/77.30/\underline{98.24}/96.66/\;6.80/21.41/96.09/\textbf{99.29}    & 93.20/95.92/\underline{99.61}/99.23/27.04/76.74/99.12/\textbf{99.71}  \\ [-3pt]
    &JSMA    & 34.91/65.79/\underline{84.16}/69.56/\textbf{99.02}/15.99/25.74/68.70  & 83.33/94.13/\underline{97.07}/93.44/\textbf{99.51}/67.52/80.55/92.53  \\   [-3pt]
    & CW  & 13.68/73.89/\underline{78.72}/11.51/\textbf{96.63}/\; 6.46/17.62/59.63     & 72.94/95.52/\underline{96.68}/66.82/\textbf{98.38}/53.11/79.81/91.12   \\   [-3pt]
      & DeepFool    &  15.55/65.93/\underline{82.18}/18.16/\textbf{98.27}/\;6.43/17.47/58.56   & 74.15/94.25/\underline{96.87}/73.99/\textbf{98.80}/54.08/79.11/90.82   \\  
     \bottomrule
    \end{tabular}
\end{table}

From the Table~\ref{table:attack results AD}, one can observe that ADCP has the best performance $11$ times out of $25$ and is the second best $8$ times out of $25$. More importantly, for the rest cases where other methods have better precision, our results are not much behind. Noted that although the critical detection paths selected by FGSM are used for the detection of AD samples, the results generalize well over non-FGSM adversarial inputs. 
Compared with LID that also 
uses only FGSM to train the detector, our results are significantly better. 

\paragraph{OOD/NS Detection} Similar to existing works~\cite{Mahalanobis,ODIN,OODL,zhao2022uniform}, we apply the same training procedure for both OOD and NS detection tasks. For all these approaches, we use the same settings as the original works for a fair comparison. The experimental results of our approach and the compared methods are shown in Table~\ref{table:OOD results OOD} (OOD detection) and Table~\ref{table:NS results OOD} (NS detection). 
\begin{table}[h]
    \centering
    \caption{Comparison of our approach with MemAE, MS, ODIN, MD, ELO and ADCL for different OOD on various models and datasets. Top 2 are highlighted with \textbf{bold} and \underline{underline}.}
    \label{table:OOD results OOD}
    \scriptsize
    \addtolength{\tabcolsep}{-2pt}
    \begin{tabular}{cccc}
        \toprule
        \multirow{2}{*}{Model} & \multirow{2}{*}{OOD}   & TPR at 95\% TNR $\uparrow$     & AUROC $\uparrow$    \\ \cline{3-4}
        &  & \multicolumn{2}{c}{MemAE / MS / ODIN / MD / ELO / ADCL / ADCP } \\
        \hline
         \multirow{2}{*}{\tabincell{c}{LeNet\\ (MNIST)} }     & F-MNIST  & 82.20/97.90/99.59/95.60/99.16/\underline{99.76}/\textbf{99.99}    &  93.89/99.26/99.81/98.81/99.77/\underline{99.85}/\textbf{99.97} \\  
                        & Omniglot   &  \textbf{100.0}/97.03/\underline{99.86}/93.62/\textbf{100.0}/\textbf{100.0}/\textbf{100.0}   & \textbf{100.0}/98.82/ 99.81/98.48/\textbf{100.0}/\underline{99.99}/\textbf{100.0}  \\  
        \midrule  
       \multirow{3}{*}{\tabincell{c}{VGG\\ (CIFAR-10)} }    &TinyIm    & 32.90/36.14/55.76/46.61/84.96/\underline{90.35}/\textbf{98.21}   & 78.03/89.79/92.79/90.68/96.51/\underline{98.23}/\textbf{99.33} \\  
                        & LSUN    & 38.12/38.82/63.41/56.92/92.13/\underline{95.77}/\textbf{99.60}    & 81.98/90.90/94.58/93.05/98.17/\underline{99.33}/\textbf{99.71}  \\  
                        & iSUN    & 30.15/39.06/63.65/54.90/89.39/\underline{95.96}/\textbf{99.34}  & 75.33/90.89/94.53/92.61/97.82/\underline{99.08}/\textbf{99.59}  \\  
                        & SVHN    & \;1.62/27.89/43.58/23.28/\underline{87.77}/73.26/\textbf{90.69}  & 28.15/89.01/91.92/85.56/\underline{97.22}/96.71/\textbf{97.82} \\  
        \midrule
       \multirow{3}{*}{\tabincell{c}{ResNet\\ (CIFAR-10)} }   &TinyIm    & 29.03/30.32/48.41/15.52/\underline{95.08}/94.97/\textbf{99.33}  &   78.69/87.13/90.58/79.46/\underline{98.93}/98.81/\textbf{99.76} \\  
                        & LSUN    & 31.18/35.47/67.04/22.84/96.21/\underline{97.73}/\textbf{99.23}    & 83.17/89.66/94.72/85.22/99.08/\underline{99.37}/\textbf{99.82}   \\  
                        & iSUN   & 24.39/34.25/63.47/22.45/93.61/\underline{96.09}/\textbf{98.89}    & 78.21/89.22/94.10/84.41/98.65/\underline{99.15}/\textbf{99.74}  \\  
                        & SVHN    &  \;1.35/38.71/67.27/24.77/85.29/\underline{86.99}/\textbf{91.16}     & 23.74/89.74/93.80/82.19/96.78/\underline{97.38}/\textbf{97.96}   \\ 
       \midrule

    \multirow{3}{*}{\tabincell{c}{VGG\\ (SVHN)} }     & TinyIm    &   43.93/78.69/88.79/76.37/92.15/\underline{95.57}/\textbf{99.98}   & 90.21/96.92/97.99/96.56/98.23/\underline{99.15}/\textbf{99.95} \\  
                        & LSUN    &  47.73/78.27/88.13/78.42/93.71/\underline{96.29}/\textbf{99.96}  & 92.16/96.80/97.83/96.71/98.40/\underline{99.32}/\textbf{99.94}  \\  
                        & iSUN   &   44.15/81.48/90.83/79.82/93.65/\underline{96.90}/\textbf{99.92}   &   90.08/97.26/98.27/96.95/98.48/\underline{99.35}/\textbf{99.93}    \\  
                        & CIFAR-10  & 33.00/78.13/88.87/76.39/72.24/\underline{90.19}/\textbf{99.56}  &  83.47/97.67/97.91/85.69/94.67/\underline{98.28}/\textbf{99.75}    \\  
         \midrule

  \multirow{3}{*}{\tabincell{c}{ResNet\\ (SVHN)} }    & TinyIm      &  65.45/75.38/85.39/63.93/93.85/\underline{96.64}/\textbf{99.88}    &  92.14/96.21/97.17/94.37/98.65/\underline{99.17}/\textbf{99.84}    \\  
                        & LSUN    & 71.54/72.67/83.02/58.96/96.34/\underline{97.46}/\textbf{99.94}  & 94.30/95.85/96.72/93.64/99.09/\underline{99.33}/\textbf{99.86}   \\ 
                       & iSUN      & 63.26/75.65/86.39/63.04/95.80/\underline{97.57}/\textbf{99.90}  &  92.25/96.28/97.30/94.20/99.01/\underline{99.35}/\textbf{99.86}   \\  
                        & CIFAR-10    &  43.24/74.12/85.12/63.91/80.38/\underline{89.70}/\textbf{97.90}    & 85.39/96.04/97.07/94.47/95.41/\underline{97.77}/\textbf{99.30}   \\ 
         \bottomrule
    \end{tabular}
\end{table}

\begin{table}[h]
    \centering
    \caption{Comparison of our approach with MemAE, MS, ODIN, MD, ELO and ADCL for different NS on various models and datasets. Top 2 are highlighted with \textbf{bold} and \underline{underline}.}
    \label{table:NS results OOD}
    \scriptsize
    \addtolength{\tabcolsep}{-2pt}
    \begin{tabular}{cccc}
        \toprule
        \multirow{2}{*}{Model} & \multirow{2}{*}{NS}   & TPR at 95\% TNR $\uparrow$     & AUROC $\uparrow$    \\ \cline{3-4}
        &  & \multicolumn{2}{c}{MemAE / MS / ODIN / MD / ELO / ADCL / ADCP } \\
       \midrule
         \multirow{2}{*}{\tabincell{c}{LeNet\\ (MNIST)} }     & Gaussian  & \textbf{100.0}/\textbf{100.0}/\textbf{100.0}/\textbf{100.0}/\textbf{100.0}/\textbf{100.0}/\textbf{100.0}    &  \textbf{100.0}/99.58/\textbf{100.0}/\underline{99.77}/\textbf{100.0}/\textbf{100.0}/\textbf{100.0}   \\  
                        & Uniform    & \textbf{100.0}/\underline{99.27}/\textbf{100.0}/99.25/\textbf{100.0}/\textbf{100.0}/\textbf{100.0}    &  \textbf{100.0}/98.31/\underline{99.92}/98.85/\textbf{100.0}/\textbf{100.0}/\textbf{100.0}   \\  
                        & FoolIm    & \textbf{100.0}/\;0.0\; /\;0.0\; /\;2.07/\textbf{100.0}/\textbf{100.0}/\underline{98.77}   &  \textbf{100.0}/75.56/86.73/68.61/\textbf{100.0}/\underline{99.80}/99.62   \\
     \midrule
       \multirow{2}{*}{\tabincell{c}{VGG\\ (CIFAR-10)} }    & Gaussian  & \textbf{100.0}/\;7.75/58.42/\underline{97.43}/\textbf{100.0}/\textbf{100.0}/\textbf{100.0}    &  \textbf{100.0}/90.26/95.21/\underline{98.83}/\textbf{100.0}/\textbf{100.0}/\textbf{100.0}    \\  
                        & Uniform     & \textbf{100.0}/49.56/87.79/\underline{99.81}/\textbf{100.0}/\textbf{100.0}/\textbf{100.0}   &  \underline{99.91}/94.72/96.78/99.28/\textbf{100.0}/\textbf{100.0}/\textbf{100.0}    \\  
                        & FoolIm   &  77.55/\;0.0\; /\;0.0\; /\;0.0\; /\textbf{100.0}/\underline{99.54}/\textbf{100.0} 
                        &  \;7.05/71.60/79.53/71.82/\textbf{100.0}/98.85/\underline{99.89}   \\
      \midrule
       \multirow{2}{*}{\tabincell{c}{ResNet\\ (CIFAR-10)} }   & Gaussian    & \textbf{100.0}/30.22/\underline{58.51}/33.67/\textbf{100.0}/\textbf{100.0}/\textbf{100.0}   &  \underline{99.98}/89.08/93.90/89.87/\textbf{100.0}/99.77/\textbf{100.0}   \\  
                        & Uniform   & \textbf{100.0}/24.56/\underline{53.40}/23.48/\textbf{100.0}/\textbf{100.0}/\textbf{100.0}      & 99.68/87.96/93.38/89.13/\textbf{100.0}/\underline{99.83}/\textbf{100.0}   \\  
                        & FoolIm   & \textbf{100.0}/\;0.0\; /\;0.0\; /\underline{16.83}/\textbf{100.0}/\textbf{100.0}/\textbf{100.0}     &  \textbf{100.0}/72.84/82.74/76.65/\textbf{100.0}/\underline{99.78}/\textbf{100.0}   \\
   \midrule

    \multirow{2}{*}{\tabincell{c}{VGG\\ (SVHN)} }    &  Gaussian   & \textbf{100.0}/84.76/91.95/80.21/\textbf{100.0}/\underline{99.85}/\textbf{100.0}      & 99.73/97.67/98.49/97.00/\underline{99.94}/99.81/\textbf{100.0}   \\  
                        & Uniform   &  47.31/90.49/96.35/83.88/\textbf{100.0}/\underline{99.66}/\textbf{100.0}  & 96.38/98.35/99.11/97.30/\underline{99.85}/99.77/\textbf{99.99}   \\  
                        & FoolIm    &  \underline{99.88}/\;0.0\; /\;0.0\; /\;0.0\; /\textbf{100.0}/100.0/\textbf{100.0}   & 99.26/16.65/19.45/37.91/\underline{99.89}/99.01/\textbf{99.96}    \\
   \midrule

  \multirow{2}{*}{\tabincell{c}{ResNet\\ (SVHN)} }    & Gaussian  & \textbf{100.0}/83.52/\underline{93.28}/66.61/\textbf{100.0}/\textbf{100.0}/\textbf{100.0}   & 99.81/97.28/98.32/95.31/\textbf{99.99}/\underline{99.94}/\textbf{99.99}   \\  
                        & Uniform     &  91.76/82.68/93.08/62.76/\textbf{100.0}/\underline{99.99}/\textbf{100.0}     & 98.28/97.28/98.32/94.89/\underline{99.93}/99.90/\textbf{99.96}   \\  
                        & FoolIm    & \textbf{100.0}/\;0.0\; /\;0.0\; /\;1.30/\textbf{100.0}/\underline{98.78}/\textbf{100.0}   &  \underline{99.71}/42.09/46.74 /53.18/\textbf{99.99}/98.90/\textbf{99.99}  \\ 
    \bottomrule
    \end{tabular}
\end{table}
From the Table~\ref{table:OOD results OOD} and Table~\ref{table:NS results OOD}, we can observe that the performance of our approach 
is mostly better than other methods for OOD and NS detection task. For the OOD detection tasks (Table~\ref{table:OOD results OOD}),  our method has the best precision in all OOD cases. For the NS detection tasks (Table~\ref{table:NS results OOD}), our method has the best precision for detection of NS anomalies in most cases. Even for the some cases when ELO is better, the percentage difference is minor.

\begin{center}
\fcolorbox{black}{gray!10}{\parbox{.95\linewidth}{\textbf{Answer to RQ2:} Compared with the state-of-the-art traditional anomaly detection methods, our approach provides more stable detection performance for AD and superior performance for OOD and NS data.}}
\end{center}

\subsection{RQ3: Comparison with Other Path Extraction Methods}
We answer \textbf{RQ3} by comparing our approach with two other path extraction based approaches, Neuron Path Coverage (NPC)~\cite{xie2022npc} and Effective Path (EffPath)~\cite{qiu2019adversarial}. We first follow their approaches to obtain critical `paths' (their `path' contains several neurons for each layer) from the training set, and then merge their `paths' for each class to obtain the final critical decision `paths'.  We define these two critical decision `paths' as Path$_{NPC}$ and Path$_{EffPath}$, respectively. 
Note that their `paths' are defined differently from ours, and for a fair comparison, we randomly extract $21$ paths (in our definition) where one neuron per layer from Path$_{NPC}$ and Path$_{EffPath}$, and denoted them as Path$_{NPC}^{*}$ and Path$_{EffPath}^{*}$. We show two experimental results of the EffPath method: the comparison with their original detection method using the similarity calculation, denoted as EffPath and the comparison using our approach by calculating scores using SVDD and Path$_{EffPath}^{*}$, donated as EffPath$^{*}$. For NPC, since 
their work does not apply effective paths for anomaly detection, the result is obtained by using our detection approach i.e., SVDD and Path$_{NPC}^{*}$. These experimental results for the detection of AD, OOD and NS anomalies are shown in Table~\ref{table:attack results path}, Table~\ref{table:OOD results path} and Table~\ref{table:NS results path}, respectively.

\begin{table}[h] 
    \caption{Comparison of our approach with NPC and EffPath for different attack on various models and datasets. The highest are highlighted with \textbf{bold}.}\label{table:attack results path}
    \scriptsize
    \centering
    \begin{tabular}{@{\hskip -0.01cm}c@{\hskip -0.01cm}ccc}
        \toprule
         \multirow{2}{*}{Model} & \multirow{2}{*}{Attack}    & TPR at 95\% TNR $\uparrow$    & AUROC $\uparrow$ \\ \cline{3-4}
          &  & \multicolumn{2}{c}{ NPC / EffPath / EffPath$^{*}$ / ADCP  }  \\
        \midrule
        \multirow{3}{*}{{\tabincell{c}{LeNet\\ (MNIST)} } }    
         & FGSM    &  \textbf{97.13}/64.14/95.10/93.99     & \textbf{99.40}/92.84/99.14/98.99   \\ 
         & PGD    &  45.39/35.59/46.21/\textbf{76.19}   & 88.49/83.72/90.37/\textbf{97.76}  \\ 
        &JSMA    & 92.60/89.77/92.91/\textbf{94.12}  & 98.56/98.01/98.57/\textbf{98.81} \\ 
        & CW  & 67.32/77.34/68.36/\textbf{82.82}     & 95.47/96.08/95.39/\textbf{97.24}   \\ 
        & DeepFool    & 71.84/76.70/72.85/\textbf{85.43}  & 95.98/95.83/95.82/\textbf{97.44}   \\     \midrule
   
    \multirow{3}{*}{{\tabincell{c}{VGG\\ (CIFAR-10)} } }  
    & FGSM    &  55.26/44.82/59.00/\textbf{99.65}   & 92.27/81.74/93.64/\textbf{99.56}   \\ 
    & PGD  &  40.64/71.20/45.96/\textbf{82.34}     & 86.94/94.75/89.32/\textbf{96.57}   \\  
    &JSMA    & 95.92/93.47/95.66/\textbf{97.46}  & 98.84/98.71/98.95/\textbf{99.15} \\  
    & CW  & 90.10/77.71/88.07/\textbf{94.92}     & 97.51/94.62/97.34/\textbf{98.97} \\ 
    & DeepFool    &  85.01/72.58/82.80/\textbf{92.42}   & 96.55/92.62/96.26/\textbf{98.54} \\
     \midrule
     
     \multirow{3}{*}{{\tabincell{c}{ResNet\\ (CIFAR-10)} } }  
         & FGSM    &  99.91/\; 8.28/99.95/\textbf{100.0}   & 99.73/72.91/99.83/\textbf{100.0}  \\ 
        & PGD  & 92.31/29.11/93.25/\textbf{99.53}  & 98.16/79.09/98.34/\textbf{99.76}  \\  
     &JSMA    & 89.31/79.89/88.91/\textbf{94.15}   &96.93/95.86/96.95/\textbf{98.42}  \\  
     & CW  & 34.54/38.59/33.24/\textbf{85.04}     & 85.92/86.72/85.58/\textbf{97.63}  \\  
    & DeepFool    & 33.94/40.64/33.13/\textbf{84.66}   & 85.81/87.57/85.37/\textbf{97.53}  \\  
  \midrule
     
     \multirow{3}{*}{{\tabincell{c}{VGG\\ (SVHN)} } }  
         & FGSM    &  96.51/68.29/98.05/\textbf{100.0}   & 99.38/92.26/99.43/\textbf{99.93}   \\ 
       & PGD   & 74.84/50.16/76.36/\textbf{89.75}  & 94.46/84.42/95.00/\textbf{97.88}  \\   
     &JSMA    & 79.71/88.44/82.36/\textbf{95.34}   & 95.68/97.08/96.21/\textbf{98.67} \\  
     & CW  & 86.28/96.64/90.23/\textbf{99.70}     & 97.72/98.39/98.22/\textbf{99.55}  \\  
     & DeepFool    &  54.08/63.90/57.55/\textbf{81.50}    & 87.23/88.87/88.11/\textbf{94.48}   \\  
   \midrule
     
    \multirow{3}{*}{{\tabincell{c}{ResNet\\ (SVHN)} } }  
        & FGSM    &  100.0/61.19/100.0/\textbf{100.0}  & 99.81/91.16/99.82/\textbf{99.90}   \\    
        & PGD  & 93.32/43.89/91.30/\textbf{99.29}  & 98.81/81.29/98.47/\textbf{99.71} \\  
    &JSMA    & 22.39/64.35/22.82/\textbf{68.70}  & 79.11//97.72/79.12/\textbf{92.53} \\  
    & CW  & 16.86/52.03/17.33/\textbf{59.63}     & 79.44/88.79/79.67/\textbf{91.12}   \\   
      & DeepFool    &  16.98/52.85/17.16/\textbf{58.56}   & 78.93/89.54/78.99/\textbf{90.82}  \\  
     \bottomrule
    \end{tabular}
\end{table}

\begin{table}[h]
    \centering
    \caption{Comparison of our approach with NPC and  EffPath for different OOD on various models and datasets. The highest are highlighted with \textbf{bold}.}
    \label{table:OOD results path}
    \scriptsize
    \addtolength{\tabcolsep}{-2pt}
    \begin{tabular}{cccc}
        \toprule
        \multirow{2}{*}{Model} & \multirow{2}{*}{OOD}   & TPR at 95\% TNR $\uparrow$     & AUROC $\uparrow$    \\ \cline{3-4}
        &  & \multicolumn{2}{c}{ NPC / EffPath / EffPath$^{*}$ / ADCP } \\
        \hline
         \multirow{2}{*}{\tabincell{c}{LeNet\\ (MNIST)} }     & F-MNIST  & 98.56/88.46/99.03/\textbf{99.99}    &  99.51/97.61/99.65/\textbf{99.97} \\ 
                        & Omniglot   &  99.61/64.38/100.0/\textbf{100.0}   & 99.91/94.83/100.0/\textbf{100.0}\\
        \midrule

       \multirow{3}{*}{\tabincell{c}{VGG\\ (CIFAR-10)} }    &TinyIm    & 81.69/53.60/80.53/\textbf{98.21}   & 95.76/83.01/95.94/\textbf{99.33} \\ 
                        & LSUN    & 90.16/60.66/89.73/\textbf{99.60}    & 98.03/87.04/98.19/\textbf{99.71}  \\  
                        & iSUN    & 90.90/50.01/89.89/\textbf{99.34}  & 97.67/86.22/97.86/\textbf{99.59}  \\  
                        & SVHN    & 65.00/26.05/68.00/\textbf{90.69}  & 94.73/78.06/95.36/\textbf{97.82} \\
        \midrule
       \multirow{3}{*}{\tabincell{c}{ResNet\\ (CIFAR-10)} }   &TinyIm    & 87.83/44.90/87.98/\textbf{99.33}  &   97.70/85.41/97.77/\textbf{99.76} \\  
                        & LSUN    & 93.19/54.41/93.35/\textbf{99.23}    & 98.66/90.33/98.67/\textbf{99.82}   \\ 
                        & iSUN   & 90.48/49.93/90.60/\textbf{98.89}    & 98.16/88.40/98.24/\textbf{99.74}  \\ 
                        & SVHN    &  89.17/31.27/88.76/\textbf{91.16}     & 97.31/79.18/97.22/\textbf{97.96}   \\
       \midrule

    \multirow{3}{*}{\tabincell{c}{VGG\\ (SVHN)} }     & TinyIm    &  97.03/71.72/96.76/\textbf{99.98}   & 99.44/88.89/99.42/\textbf{99.95} \\ 
                        & LSUN    &  97.81/74.10/97.81/\textbf{99.96}  & 99.57/90.97/99.60/\textbf{99.94}  \\ 
                        & iSUN   &   97.95/77.34/97.86/\textbf{99.92}   &  99.55/91.58/99.56/\textbf{99.93}    \\ 
                        & CIFAR-10  & 92.70/74.43/92.29/\textbf{99.56}  &  98.79/90.26/98.70/\textbf{99.75}    \\
         \midrule

  \multirow{3}{*}{\tabincell{c}{ResNet\\ (SVHN)} }    & TinyIm      &  97.81/71.38/98.08/\textbf{99.88}    & 99.41/88.53/99.42/\textbf{99.84}    \\ 
                        & LSUN    & 98.86/70.37/99.04/\textbf{99.94}  & 99.62/88.81/99.64/\textbf{99.86}   \\ 
                       & iSUN      & 98.71/71.76/98.90/\textbf{99.90}  &  99.57/88.81/99.59/\textbf{99.86}   \\ 
                        & CIFAR-10    & 91.87/71.27/92.65/\textbf{97.90}    & 98.11/88.54/98.27/\textbf{99.30}   \\
         \bottomrule
    \end{tabular}
\end{table}

\begin{table}[h]
    \centering
    \caption{Comparison of our approach with NPC and EffPath for different NS on various models and datasets. The highest are highlighted with \textbf{bold}.}
    \label{table:NS results path}
    \scriptsize
    \addtolength{\tabcolsep}{-2pt}
    \begin{tabular}{cccc}
        \toprule
        \multirow{2}{*}{Model} & \multirow{2}{*}{NS}   & TPR at 95\% TNR $\uparrow$     & AUROC $\uparrow$    \\ \cline{3-4}
        &  & \multicolumn{2}{c}{ NPC / EffPath / EffPath$^{*}$ / ADCP } \\
       \midrule
         \multirow{2}{*}{\tabincell{c}{LeNet\\ (MNIST)} }     & Gaussian  & \textbf{100.0}/71.40/\textbf{100.0}/\textbf{100.0}    & \textbf{100.0}/95.68/\textbf{100.0}/\textbf{100.0}   \\ 
                        & Uniform    & \textbf{100.0}/11.15/\textbf{100.0}/\textbf{100.0}    &  \textbf{100.0}/69.53/\textbf{100.0}/\textbf{100.0}   \\ 
                        & FoolIm    & \textbf{100.0}/\; 8.47/\textbf{100.0}/98.77   & \textbf{99.97}/54.37/99.95/99.62   \\
     \midrule
       \multirow{2}{*}{\tabincell{c}{VGG\\ (CIFAR-10)} }    & Gaussian  & 79.47/\textbf{100.0}/\textbf{100.0}/\textbf{100.0}    & 99.99/95.71/\textbf{100.0}/\textbf{100.0}    \\ 
                        & Uniform     & \textbf{100.0}/98.21/\textbf{100.0}/\textbf{100.0}   &  99.99/99.25/\textbf{100.0}/\textbf{100.0}    \\ 
                        & FoolIm   &  \textbf{100.0}/\;0.04/\textbf{100.0}/\textbf{100.0} 
                        &  99.38/48.39/99.60/\textbf{99.89}   \\
      \midrule
       \multirow{2}{*}{\tabincell{c}{ResNet\\ (CIFAR-10)} }   & Gaussian    & \textbf{100.0}/82.18/\textbf{100.0}/\textbf{100.0}   & \textbf{100.0}/93.46/\textbf{100.0}/\textbf{100.0}   \\  
                        & Uniform   & \textbf{100.0}/63.26/\textbf{100.0}/\textbf{100.0}      & \textbf{100.0}/93.22/\textbf{100.0}/\textbf{100.0}     \\  
                        & FoolIm   & \textbf{100.0}/\;0.08/\textbf{100.0}/\textbf{100.0}     &  \textbf{100.0}/41.66/\textbf{100.0}/\textbf{100.0}   \\
   \midrule

    \multirow{2}{*}{\tabincell{c}{VGG\\ (SVHN)} }    &  Gaussian   & \textbf{100.0}/59.15/\textbf{100.0}/\textbf{100.0}      & 99.99/89.01/99.99/\textbf{100.0}   \\  
                        & Uniform   &  \textbf{100.0}/69.58/\textbf{100.0}/\textbf{100.0}  & 99.98/90.89/99.97/\textbf{99.99}   \\  
                        & FoolIm    &  \textbf{100.0}/\;2.99/\textbf{100.0}/\textbf{100.0}   & 99.88/48.54/99.90/\textbf{99.96}    \\
   \midrule

  \multirow{2}{*}{\tabincell{c}{ResNet\\ (SVHN)} }    & Gaussian  & \textbf{100.0}/88.92/\textbf{100.0}/\textbf{100.0}   & 99.97/98.03/99.97/\textbf{99.99}   \\ 
                        & Uniform     & \textbf{100.0}/89.14/\textbf{100.0}/\textbf{100.0}     & 99.91/97.96/99.91/\textbf{99.96}   \\  
                        & FoolIm    & \textbf{100.0}/\;0.18/\textbf{100.0}/\textbf{100.0}   &  99.96/42.98/99.94/\textbf{99.99}  \\
    \bottomrule
    \end{tabular}
\end{table}

From Table~\ref{table:attack results path}, Table~\ref{table:OOD results path} and Table~\ref{table:NS results path}, one can observe that when applied to the anomaly detection task, paths extracted using our approach are obviously superior to paths extracted using other approaches. In addition, the EffPath detection approach that uses the measure of similarity to detect anomaly samples, are not effective for OOD detection and invalid for Fooling images detection (see the results of Effpath in Table~\ref{table:OOD results path} and Table~\ref{table:NS results path}). This is because the paths extracted in their approach are obtained according to the predicted probability. When the predicted probability of an anomaly sample is close to the normal samples, the path extracted is the same as the path extracted from normal samples, which is not conducive in the anomaly detection task.

\begin{center}
\fcolorbox{black}{gray!10}{\parbox{.95\linewidth}{\textbf{Answer to RQ3:} Our \emph{critical detection paths} outperform other \emph{critical decision paths} approaches for the anomaly detection task.}}
\end{center}

\subsection{The Setting of Hyper-parameters}
We answer \textbf{RQ4} by discussing the optimal set up of the two input parameters in the Algorithm~\ref{alg:1}: the number of mutations $n$ and the number of critical detection paths $m$. 

\begin{figure}[h]
  \centering
  \includegraphics[scale=0.45]{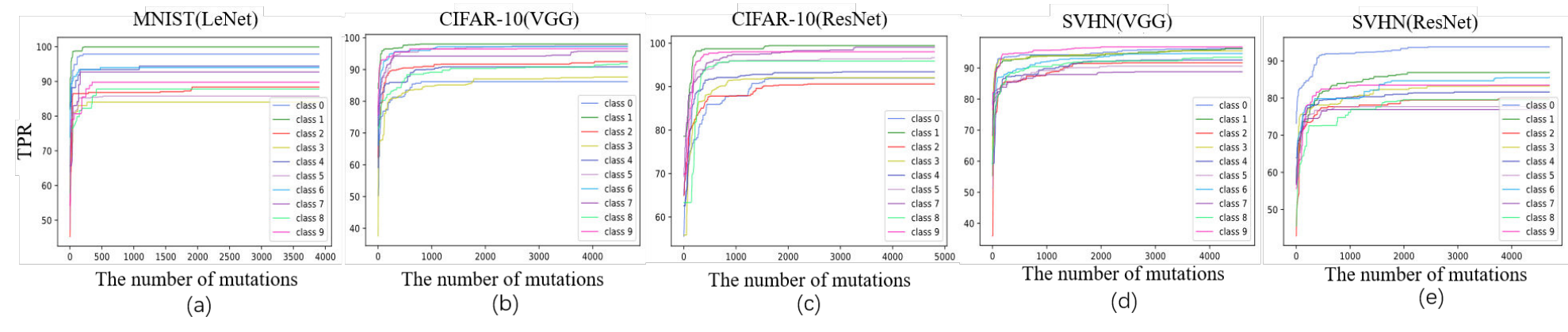}
  \caption{The TPR corresponding to the mutation number of ADCP on different models for adversarial attack in each class. }\label{fig:AD_mutations}
\end{figure}

\begin{figure}[h]
  \centering
  \includegraphics[scale=0.44]{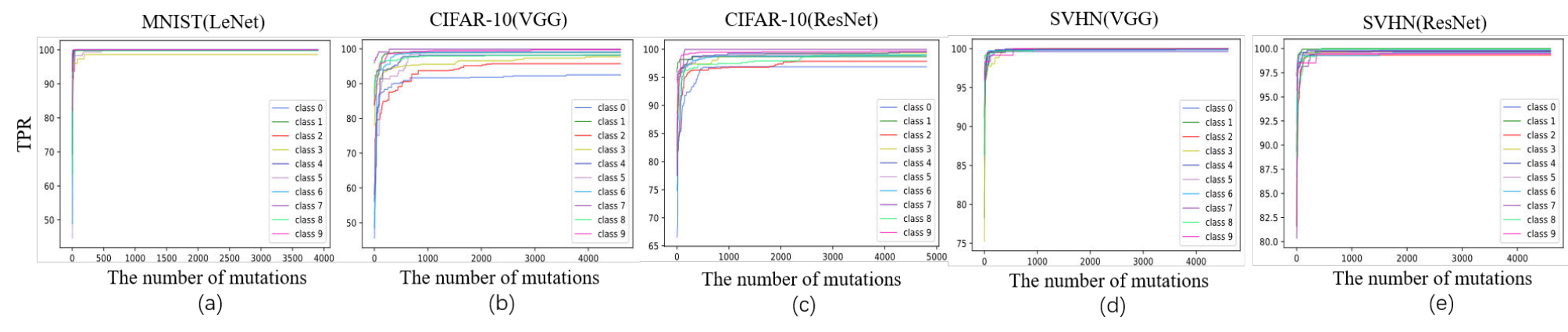}
  \caption{The TPR corresponding to the mutation number of ADCP on different models for OOD in each class.}\label{fig:OOD_mutations}
\end{figure}

\paragraph{The Number of Mutations}
We conduct an experiment to study the relationship between the number of mutations applied and the TPRs during the path generation, in order to find a reasonably small $n$ which still guarantees close to optimal performance. To this end, we assign parameter $n=5000$, and show the variation of TPRs for different models (see Figure~\ref{fig:AD_mutations} for AD detection and Figure~\ref{fig:OOD_mutations} for OOD detection).

From the Figure~\ref{fig:AD_mutations} and Figure~\ref{fig:OOD_mutations}, we can draw three conclusions. First, for all models, the TPRs become stable within $5000$ iterations, meaning that setting $n=5000$ ensures that the identified critical detection path is representative. Second, the TPR tends to stabilize faster on OOD than AD for a DNN model. Third, the TPRs of critical detection paths for different classes of a model are often different.

\begin{figure}[h]
  \centering
  \includegraphics[scale=0.45]{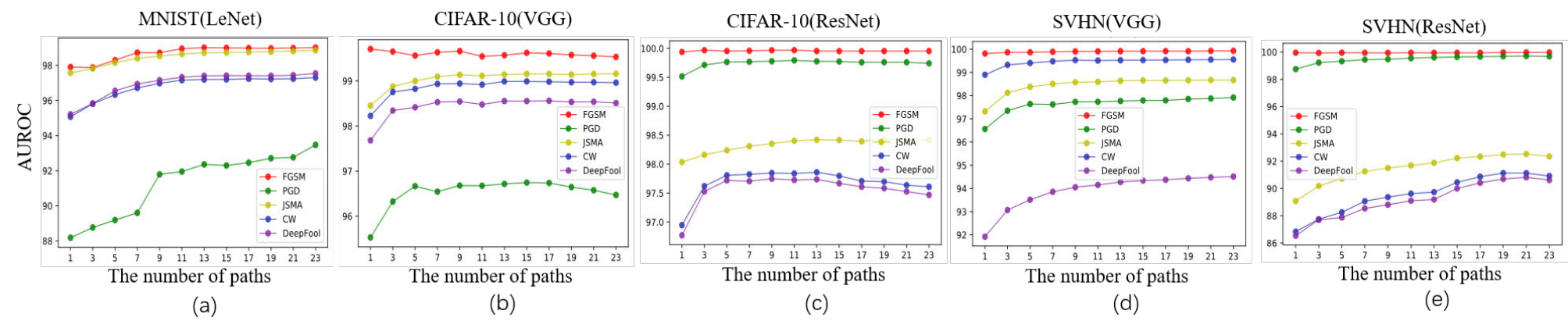}
  \caption{The AUROC corresponding to the number of paths on different models for different adversarial attacks. }\label{fig:AD_paths}
\end{figure}

\begin{figure}[h]
  \centering
  \includegraphics[scale=0.45]{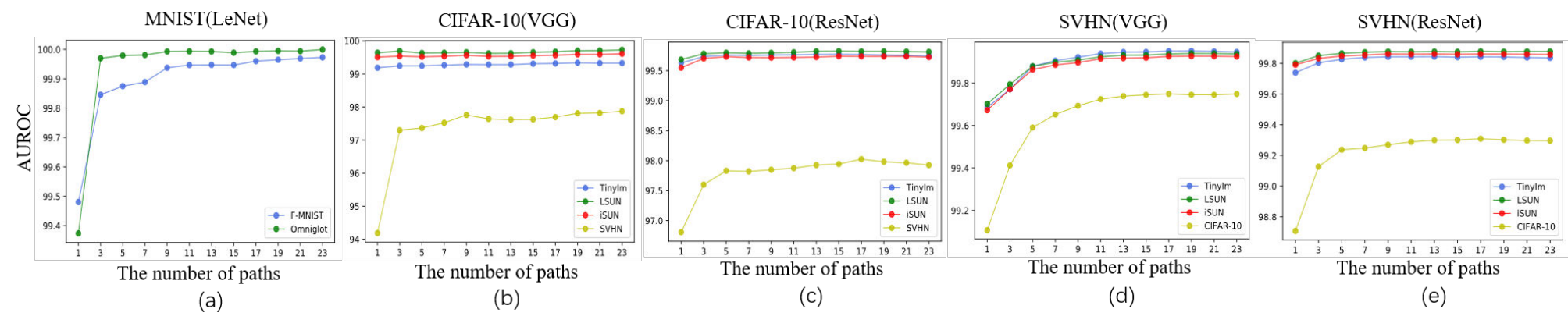}
  \caption{The AUROC corresponding to the number of paths on different models for different OOD data. }\label{fig:OOD_paths}
\end{figure}

\paragraph{The Number of Paths} 
Plainly, applying more critical detection paths often yields better performance with the cost of more execution time. Hence we study the relationship between the number of paths and the detection results. The results for AD an OOD/NS detection are shown in Figure~\ref{fig:AD_paths} and Figure~\ref{fig:OOD_paths}, respectively.

From the Figure~\ref{fig:AD_paths} and  Figure~\ref{fig:OOD_paths}, we can observe that as the number of critical paths increases, the detection results first gradually increase to a certain level, and then they tend to remain stable. Especially when the number of critical detection paths increases from 1 to 3, the detection results of most anomaly samples increase significantly. 
After we reach the stage of $3$ paths, the performance increase tends to slow down. From the current results, the ensemble from $21$ critical detection paths seem to guarantee reasonably good detection rates for all anomaly types in the experiment.

\begin{center}
\fcolorbox{black}{gray!10}{\parbox{.95\linewidth}{\textbf{Answer to RQ4:} The parameter setup in the experiments---the mutation number set to 5000 and the path number set to 21, is reasonable.}}
\end{center}

\subsection{Analysis of Each Individual Critical Detection Path}
\begin{figure}[h]
  \centering
  \includegraphics[scale=0.45]{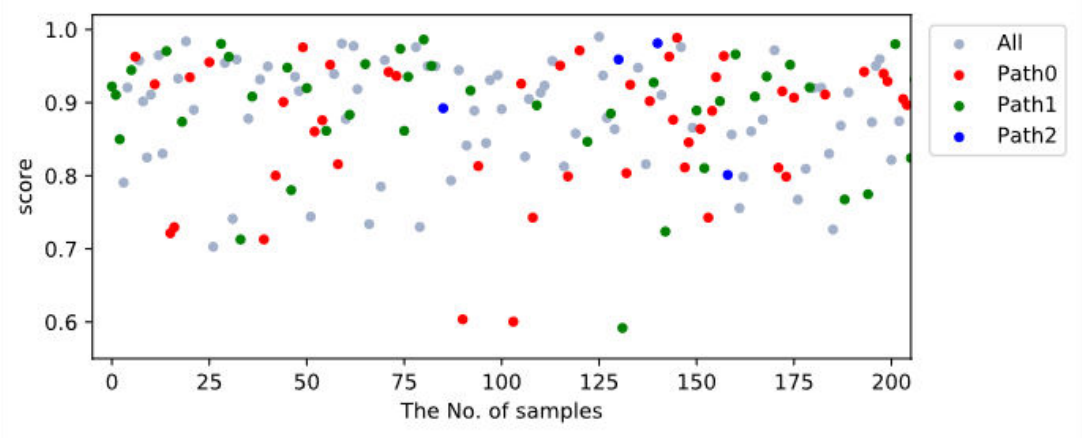}
   \vspace{-2mm}
  \caption{The visualization of 205 samples correctly decided corresponding to Top 3 critical detection paths for PGD (class 3) on MNIST.}\label{fig:scores_distribution}
\end{figure}

We conduct experiments to analyze the importance of each path in the detection. Taking the LeNet model with the PGD attack as an example, we study the anomaly samples (PGD) detected using three different critical paths with the highest TPRs. The results is partially visualized (for one decision class only) in Figure~\ref{fig:scores_distribution}, which shows the indices of the TN outputs of the three critical paths. 
 In the figure, the gray points represent the PGD adversarial input that can be detected correctly by all the three paths; the red points represent the inputs that can be additionally detected by 
 path0, the green points are the inputs that can be further detected by path1, and the blue points are inputs that are further detected by path2. 

 We can observe that a majority of the PGD inputs can be detected by all three paths. However, each individual path can detect a small set of extra anomaly inputs. This indicates that 
 combining multiple paths can improve detection~performance.

\subsection{Pearson Coefficient of Critical Detection Paths for Different Anomaly Samples}
\begin{figure}[h]
  \centering
  \includegraphics[scale=0.65]{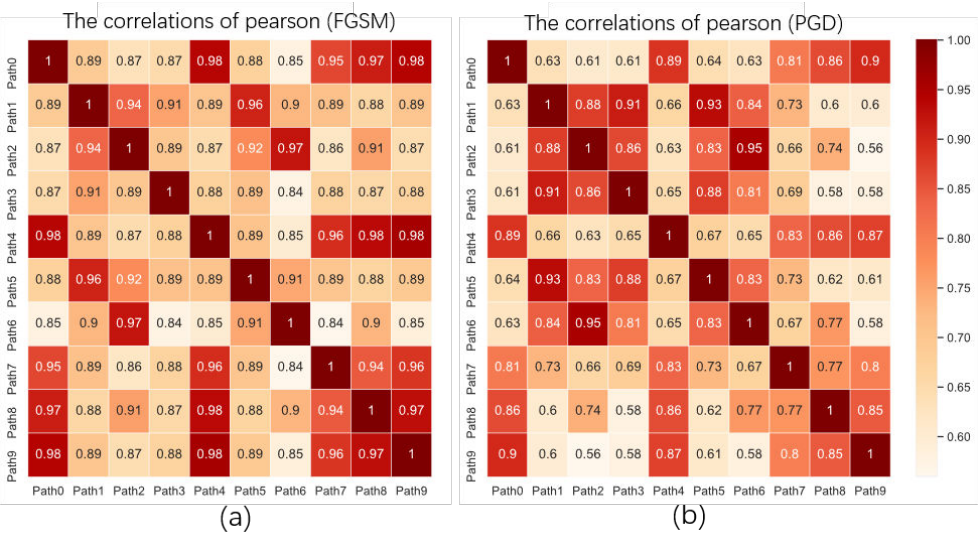}
  \vspace{-4mm}
  \caption{The correlations of pearson corresponding to Top 10 critical detection paths for FGSM and PGD on MNIST.}\label{fig:LeNet_pearson}
  \vspace{-7mm}
\end{figure}

As shown the Figure~\ref{fig:AD_paths} and Figure~\ref{fig:OOD_paths}, the magnitude of performance improvements is obviously different for different anomaly samples. Therefore, we investigate the characteristics of their corresponding critical detection paths that lead to this phenomenon. In theory, if two paths are closely correlated, i.e., with high the Pearson coefficient, the performance cannot benefit much by combining these two paths. 
Hence, we conduct experiments to show the differences of their critical detection paths (in terms of Pearson coefficient) with different anomaly samples. We select $10$ paths whose TPRs are the top $10$ among the $m$ critical detection paths and take the FGSM and PGD attacks on MNIST as example. We first calculate the scores on these $10$ critical detection paths for FGSM and PGD respectively. Then we calculate the Pearson correlation coefficients based on their scores, and the results are shown in Figure~\ref{fig:LeNet_pearson}.  
We can observe that different types anomaly inputs, such as FGSM and PGD, may induce different Pearson correlation coefficients. The Pearson correlation coefficient of PGD is generally smaller than that of FGSM. Correspondingly, combining multiple paths can significantly improve performance for PGD, but not as effective for FGSM, as shown in (a) of Figure~\ref{fig:AD_paths}.

\subsection{The Influence of Different Anomaly Samples to the Detection Performance}

\begin{figure}[h]
  \centering
  \includegraphics[scale=0.8]{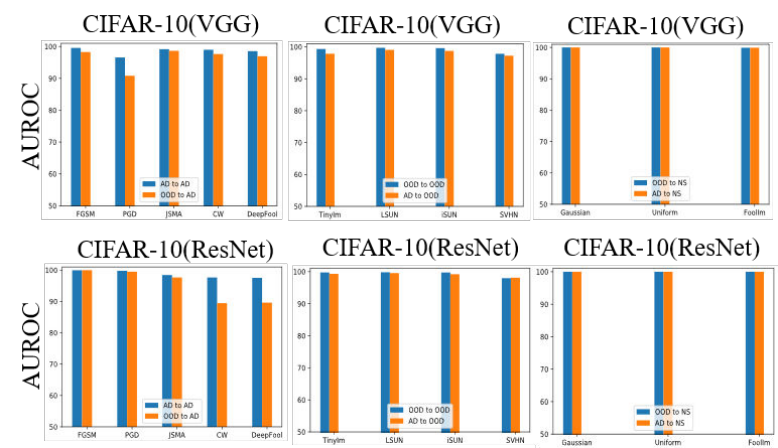}
  \vspace{-4mm}
  \caption{The AUROC corresponding to the paths selected with OOD samples applied to AD detection and the paths selected with AD applied to OOD/NS detection on CIFAR-10.}\label{fig:AD_to_OOD}
  \vspace{-4mm}
\end{figure}

As presented in Subsection~\ref{approach:extraction}, for the AD detection task, we use AD data as anomaly samples to select one critical detection path. For the OOD/NS detection task, we use OOD data as anomaly samples to select one critical detection path. 
To investigate whether this separation is necessary i.e. test the generalization of different critical detection paths in anomaly detection, we conduct experiments to show whether a path selected by using AD samples is suitable in OOD/NS detection, and whether a path selected by using 
OOD anomaly set is suitable in AD detection. We take the DNN models with in-distribution CIFAR-10 as an example and illustrate the experimental results using the evaluation of AUROC integrated with 21 critical detection paths as shown in Figure~\ref{fig:AD_to_OOD}. 

For the two charts on the left in Fig.~\ref{fig:AD_to_OOD}, a blue bar (AD-to-AD) indicates that the critical detection paths selected by the AD samples are applied to detect AD samples, and an orange bar (OOD-to-AD) indicates that the critical detection paths selected by the OOD samples are applied to detect AD samples. We can see that the performance of OOD-to-AD is slightly less than AD-to-AD for most cases, the difference is large in only a few cases. 
For the 
remaining charts, a blue bar (OOD-to-OOD/NS) indicates that the critical detection paths selected by the OOD samples are applied to detect OOD/NS samples and orange bar (AD-to-OOD/NS) indicates that the critical detection paths selected by the AD samples are applied to detect OOD/NS samples.  
We can see that the performance of AD-to-OOD/NS is basically consistent with OOD-to-OOD/NS for most cases.
In summary, the paths selected by different anomaly samples do not have a significant impact on the detection performance, even though the path selected by the specific types of anomaly samples applied to detect the corresponding anomaly sample can get better results.

\subsection{Time complexity}

In this section, we analyze the time overhead of NPC, EffPath and our approach, regarding path generation. 
Note that our approach aims to find several critical paths for each class, and we determine whether an input $x$ is anomaly based on the features of these critical paths. 
NPC and EffPath apply profiling on the entire sample set by generating a path (which may contain multiple neurons per layer) for each input. A final path is then abstracted from all individually generated paths from the inputs. For these two approaches, whether an input is anomaly is determined by comparing its path information and the abstracted path information generated from the training set.
Therefore, the time overhead of our approach is mainly the time of extracting one critical path 
with iterative mutation, and the time overhead of NPC and EffPath mainly depends on the 
size of a training set.
Here, we record the time of $5000$ mutations of our approach and the time of evaluating adversarial samples with $21$ critical paths, as shown in the Table~\ref{table:ADCP complexity}. We record the time of extracting the path for one sample and the total time of extracting the paths on the training set for NPC and EffPath, as shown in the Table~\ref{table:NPC/EffPath complexity}.

From the Table~\ref{table:ADCP complexity}, we can observe that both the number of samples and the depth (the number of layers) of the target model affect the time overhead for our approach. Taking the CIFAR-10 and SVHN as an example, we use a 16-layer VGG model with the same structure as target model, while the time overhead on SVHN is significantly higher than that on CIFAR-10 as the number of samples on SVHN is larger than that on CIFAR-10.  SVHN has $73,257$ and $26,032$ samples on the training set and test set, while CIFAR-10 has $50,000$ and $10,000$ samples on the training set and test set respectively. When we extract critical path, our approach need train SVDD models and calculate $TPR$. Therefore, the larger the number of samples, the longer the time is consumed. The deeper the target model, the more features are input the SVDD models, and the more time also are consumed to train the SVDD models. 

From the Table~\ref{table:NPC/EffPath complexity},  we can observe that the time overhead also depends on the number of samples and the depth of the target model for NPC and EffPath. Since 
we need to calculate the path information for each sample separately, the larger number of samples in the training set, the more time will be consumed for the same model. Especially, we can observe that with the increase of the number of model layers, the time overhead increases significantly for different target models. This is mainly because both methods need to calculate the path information layer by layer, and the more layers, the greater amount of computation. In addition, EffPath is slightly more efficient than NPC as EffPath sets a threshold in advance and only selects neurons which have the activation values larger than the threshold as critical neurons to form a critical path, while NPC needs to calculate the relevance of all neurons and then select the critical neurons according to the size on the relevance to form the critical path. Therefore, for EffPath, the number of neurons which 
require calculation is smaller than NPC.

\begin{table}[t] 
    \caption{Time complexity of critical operations for ADCP (Seconds).}\label{table:ADCP complexity}
    \small
    \centering
    \begin{tabular}{ccccc}
        \toprule
          \multirow{1}{*}{{\tabincell{c}{Model} } }  & Dataset   & Enumerate $5000$ times    & Evaluation  & Total \\
        \midrule
       \multirow{1}{*}{{\tabincell{c}{LeNet} } }   & MNIST   & 5740.32   & 44.01  & 5514.33 \\
     \midrule
     
      \multirow{1}{*}{{\tabincell{c}{VGG} } }   & CIFAR-10   &  9680.53  & 70.97  & 9751.50  \\
      \midrule
     
    \multirow{1}{*}{{\tabincell{c}{ResNet} } }   & CIFAR-10   & 19563.68   & 132.25  & 19695.93 \\
     
     \midrule
     
    \multirow{1}{*}{{\tabincell{c}{VGG} } }   & SVHN   &  26743.37    & 239.91  & 26983.28  \\
     \midrule
     
    \multirow{1}{*}{{\tabincell{c}{ResNet} } }  & SVHN   &  30722.56     & 259.31  & 30981.87   \\ 
     \bottomrule
    \end{tabular}
\end{table}

\begin{table}[t] 
    \caption{Time complexity of extracting the path for one sample to NPC and EffPath (Seconds).}\label{table:NPC/EffPath complexity}
    \small
    \centering
    \begin{tabular}{cccccc}
        \toprule
          \multirow{1}{*}{{\tabincell{c}{Model} } }  & Dataset   & NPC    & EffPath  & NPC (Training set)  & EffPath (Training set)\\
        \midrule
       \multirow{1}{*}{{\tabincell{c}{LeNet} } }   & MNIST   & 0.39  & 0.34  & 23400.00  & 20400.00  \\
     \midrule
     
      \multirow{1}{*}{{\tabincell{c}{VGG} } }   & CIFAR-10   &  9.43   & 5.72  & 471500.00  & 28600.00 \\
      \midrule
     
    \multirow{1}{*}{{\tabincell{c}{ResNet} } }   & CIFAR-10   & 32.25   & 24.13  & 1612500.00  & 1206500.00 \\
     
     \midrule
     
    \multirow{1}{*}{{\tabincell{c}{VGG} } }   & SVHN   &  9.38  & 5.81  & 687150.66  & 425623.17\\
     \midrule
     
    \multirow{1}{*}{{\tabincell{c}{ResNet} } }  & SVHN   &  16.06  & 11.84  & 1176507.42  &  867362.88 \\ 
     \bottomrule
    \end{tabular}
\end{table}

\section{Related Works} \label{sec:related works}

In this section we summarize the existing literature related to our work.

\subsection{Anomaly Detection}
Existing works for anomaly detection either focus on AD detection, e.g., ~\cite{kd+bu,LID,InfluenceFunction,FeatureSqueezing,vacanti2020adversarial,ma2019nic,meng2017magnet}, or OOD and NS-\Romannum{1} detection, e.g.,~\cite{baseline,ODIN,OODL,OutlierExposure,GeneralizedODIN,gong2019memorizing}, or both~\cite{Mahalanobis,zhao2022uniform,raghuram2021general,lust2022efficient}. 
Notably only one work is capable of detection all types~\cite{zhao2022uniform} with acceptable performance. All these anomaly detection methods are based on the features of one layer or several layers of the DNN model. For example, Ma et al.~\cite{ma2019nic} considered the features of each layer of the DNN model, and proposed Network Invariant Checking (NIC) for the detection of AD data. Lee et al.~\cite{Mahalanobis} calculated the Mahalanobis distance by utilizing the features of each layer of the DNN model for the detection of AD and OOD/NS-I data. Hendrycks et al.~\cite{baseline} and Liang et al.~\cite{ODIN} analyze the output of DNN models for the detection of OOD/NS-I data. Zhao et al. utilized the outputs of the discriminative layer and the logit layer of a DNN model for detecting three types of anomaly samples~\cite{zhao2022uniform}. We compared our approach with all these methods and the experimental results are shown in Section~\ref{sec:evaluations}.

The existing work has proved that the features extracted by each layer of a DNN model are different~\cite{dosovitskiy2016inverting}. Therefore, 
it is reasonable to assume that detection accuracy of the existing methods can be further improved as they only rely
on a specific layer to detect anomaly samples. For the methods that consider the features of several layers, they essentially obtain a detection result based on the features of each layer, and then integrate the results to obtain a final result. However, during the integration process, anomaly samples are usually used to tune parameters~\cite{Mahalanobis}. This is often unrealistic in real-world applications. 

Due to the above defects for the existing anomaly detection methods based on features of layers, in recent years, some works jump out the layer-based paradigm, and look for vulnerabilities in neural networks from the perspective of software testing~\cite{pei2017deepxplore,ma2018deepgauge,kim2019guiding,wang2019adversarial,Munn2018,ConcolicTesting}.
Among these methods, the methods based on the neuron coverage have got much attention~\cite{pei2017deepxplore,ma2018deepgauge,kim2019guiding}. Based on the neuron coverage, DeepPath~\cite{wang2019deeppath} initially proposes the path-driven coverage criteria, which considers the sequentially linked connections of a DNN, and the generated paths represent the information transmission in the entire DNN model.  
At present, some works have focused on the idea of path-based anomaly detection, and proposed methods to extract the decision logic paths and the adversarial perturbation paths of a DNN~model. 

\subsection{Critical Paths Extraction}
Several works have been devoted to interpreting deep neural models by extracting critical paths~\cite{yu2018distilling,wang2018interpret,qiu2019adversarial,zhang2020dynamic,xie2022npc,li2021understanding}. 
These works can be classified into two categories depending on their application: used for anomaly detection, or used for interpreting a model's decision procedure.

For the works related to anomaly detection, 
Wang et al.~\cite{wang2018interpret} used knowledge distillation~\cite{hinton2006fast} to extract critical data routing paths (CDRPs) by learning associated control gates for each layer’s output channel.  They use the information of CDRPS to detect adversarial examples.
Jiang et al.~\cite{wang2018interpret} used a similar method to extract critical paths for detecting backdoor attacks. Backdoor attacks are launched by triggers, which are well-designed patterns stuck on images, and aim to mislead DNN models to generate wrong predictions~\cite{BadNets}. Although these two methods have shown good performance in detecting adversarial samples and backdoor attacks, they do not have the merge capability, as a single channel can have different significance values for different images. As such, these methods have weak generalizability~\cite{qiu2019adversarial}. 
To this end, Qiu et al.~\cite{qiu2019adversarial} tried to extract synapse-wise effective paths for each class through network profiling, which contains the neurons and connections that have a positive effect on the prediction results. They use an effective path similarity based method to detect adversarial samples. It calculates the similarity of the effective paths between one test sample and the normal samples. Theoretically, the similarity values of normal samples is higher than that of adversarial samples. However,  when the anomaly sample is predicted as a target class with high confidence, its effective path is similar to normal ones, which makes this method invalid (as shown in our evalution in Table~\ref{table:NS results path}). Since the method of effective paths only considers the neurons with a positive effect on the prediction results, their effective paths may be incomplete and imprecise. 
Zhang et al.~\cite{zhang2020dynamic} proposed a method to extract critical paths for each class by using the idea of program slicing, with regard to the effective path, their paths contain neurons and synapses that play a positive and negative contribution in the prediction~results. 

Regarding critical paths used in interpreting model decision, recently, Xie et al.~\cite{xie2022npc} analyzed the relevance of each neuron with the prediction results by using the Layer-wise Relevance Propagation method, and the decision logic paths contain neurons with high relevance (referred to as the NPC method). Li et al.~\cite{li2021understanding} proposed a gradient-based influence layer-wise propagation method to extract critical attacking paths, and these paths contain neurons with high gradient~values.

We have made a detailed discussion regarding the limitations of aforementioned methods in Section~\ref{sec:introduction}.
In Section~\ref{sec:evaluations}, we have also compared our approach with two of them (i.e., Effective Path~\cite{qiu2019adversarial} \footnote{https://github.com/pnnl/DeepDataProfiler} and NPC~\cite{xie2022npc}) \footnote{https://github.com/ltl7155/NPC} which have the executable code available to the public.

\section{Conclusion and Future Work}
We propose a novel approach called Anomaly Detection Based on Critical Paths (ADCP) based on an evolution and mutation mechanism to extract critical detection paths.  This approach does not modify the DNN structure and it extracts meaningful path information that can be used to distinguish between normal data and anomaly data. 
Through the analysis, we also find that we can further enhance  
performance via an ensemble methodology with multiple critical detection paths.
Moreover, our method 
generalizes to a broad range of anomaly samples, including adversarial inputs (AD), out-of-distribution inputs (OOD), and noise inputs (NS), all with satisfactory performance. Finally, we have conducted extensive experiments to evaluate our approaches, showing that it outperforms state-of-the-art approaches.

One limitation is that this work focus only on image classification. As the application domains of DNN is expanding fast, it is interesting and necessary to explore whether the existing methodology can be adopted to other applications beyond image processing, such as speech recognition, natural language processing, and intrusion detection with network traffic monitoring. Moreover, the key operation of our approach as described in Algorithm~\ref{alg:1} may be regarded too expensive in some scenarios which will limit its applicability, especially for large target DNNs and large class sizes. Finally, the method that we used to extract critical path can be further improved as it lacks mathematical proof. In the future, we will explore better and more effective methods for selecting critical path, so as to further enhance the interpretability of our detection results.

\begin{acks}
*****
\end{acks}

\bibliographystyle{ACM-Reference-Format}
\bibliography{cas-refs}

\end{document}